\documentclass{article}
\usepackage{iclr2026_conference,times}
\usepackage[toc,page,header]{appendix}

\usepackage{natbib}

\usepackage{amsmath}
\usepackage{amssymb}
\usepackage{mathtools}
\usepackage{amsthm}
\newtheorem{assumption}{Assumption}
\newtheorem{lemma}{Lemma}
\newtheorem{proposition}{Proposition}
\usepackage{minitoc}

\usepackage{booktabs}

\usepackage{algcompatible}
\usepackage[most]{tcolorbox}

\usepackage{amsmath}
\usepackage{amsthm}

\usepackage{amssymb}  

\usepackage{floatflt}
\usepackage{wrapfig}
\usepackage{hyperref}
\usepackage{cleveref} 
\usepackage{url}
\newcommand{\algname}{AT-GRPO}
\usepackage[dvipsnames,svgnames]{xcolor}
\usepackage[most]{tcolorbox}
\usepackage{beramono} 
\usepackage{minted}
\usepackage{algorithm,algpseudocode,amsmath}
\usepackage[most]{tcolorbox}

\newcommand{\squishlist}{
    \begin{list}{$\bullet$}{%
        \setlength{\itemsep}{0pt}%
        \setlength{\parsep}{0pt}%
        \setlength{\topsep}{0pt}%
        \setlength{\partopsep}{0pt}%
        \setlength{\listparindent}{-2pt}%
        \setlength{\itemindent}{-5pt}%
        \setlength{\leftmargin}{1.2em}%
        \setlength{\labelwidth}{0em}%
        \setlength{\labelsep}{0.5em}%
    }
}

\newcommand{\squishend}{\end{list}}
\usepackage[normalem]{ulem}
\usepackage{array}
\newcolumntype{L}[1]{>{\raggedright\arraybackslash}p{#1}}
\newcolumntype{T}[1]{>{\centering\arraybackslash}p{#1}}  
\definecolor{myblue}{rgb}{0.9, 0.95, 1.0}
\definecolor{mygray}{rgb}{0.95, 0.95, 0.95}
\definecolor{mygreen}{rgb}{0.9, 1.0, 0.9}
\definecolor{darkblue}{rgb}{0.0, 0.0, 0.5}
\usepackage{booktabs}
\usepackage{pifont}
\newcommand{\cmark}{\ding{51}} 
\newcommand{\xmark}{\ding{55}} 
\usepackage{booktabs,multirow,makecell,xcolor,colortbl,threeparttable}
\definecolor{gold}{HTML}{FFD700}
\definecolor{silver}{HTML}{C0C0C0}
\newcommand{\sdneg}[2]{\makecell{#1\\\scriptsize(\textcolor{red!70!black}{#2})}}
\newcommand{\best}[1]{\cellcolor{gold!25}\textbf{#1}}
\newcommand{\second}[1]{\cellcolor{silver!25}#1}
\newcommand{\sd}[2]{\makecell{#1\\\scriptsize(+#2)}}
\author{Yujie Zhao\textsuperscript{1} \quad
Lanxiang Hu\textsuperscript{1} \quad
Yang Wang\textsuperscript{2} \quad
Minmin Hou\textsuperscript{2} \quad \\
\textbf{Hao Zhang}\textsuperscript{1} \quad
\textbf{Ke Ding}\textsuperscript{2} \quad
\textbf{Jishen Zhao}\textsuperscript{1} \\
\textsuperscript{1}University of California, San Diego \quad
\textsuperscript{2}Intel Corporation 
}
\newif\ifshowrevisions
\showrevisionsfalse     

\ifshowrevisions
  \newcommand{\revised}[1]{\textcolor{blue}{#1}}
\else
  \newcommand{\revised}[1]{#1}
\fi
\ifshowrevisions
  \newcommand{\revisedbegin}{\begingroup\color{blue}}
  \newcommand{\revisedend}{\endgroup}
\else
  \newcommand{\revisedbegin}{\begingroup}
  \newcommand{\revisedend}{\endgroup}
\fi
\ifshowrevisions
  \newcommand{\deleted}[1]{\textcolor{black}{\sout{#1}}}
\else
  \newcommand{\deleted}[1]{}
\fi

\ifshowrevisions
  \newenvironment{deletedblock}{%
    \begingroup
    \color{red}%
    \markoverwith{\rule[0.5ex]{2pt}{0.4pt}}%
    \ULon
  }{%
    \ULoff
    \endgroup
  }
\else
  
\fi

\newcommand{\deletedbegin}{\begin{deletedblock}}
\newcommand{\deletedend}{\end{deletedblock}}
\ifshowrevisions
  
\else
\fi
\usepackage[dvipsnames]{xcolor}
\usepackage{pifont}
\usepackage[dvipsnames]{xcolor} 
\newcommand{\good}{\textcolor{ForestGreen}{\(\checkmark\)}}
\newcommand{\bad}{\textcolor{BrickRed}{\(\times\)}}
\usepackage{fontawesome5}
\definecolor{envcolor}{HTML}{F5F7FF}   
\definecolor{envborder}{HTML}{3B82F6}  

\definecolor{toolcolor}{HTML}{EEF6FF}  
\definecolor{toolborder}{HTML}{2563EB} 

\definecolor{plancolor}{HTML}{FFF5F5}  
\definecolor{planborder}{HTML}{F87171} 
\usepackage[makeroom]{cancel}

\definecolor{testcolor}{HTML}{ECFDF5}  
\definecolor{testborder}{HTML}{22C55E} 

\title{Stronger-MAS: Multi-Agent Reinforcement
Learning for Collaborative LLMs}
\iclrfinalcopy
\begin{document}

\maketitle

\begin{abstract}
Multi-Agent System (MAS) and Reinforcement Learning (RL) 
are both 
widely adopted to improve large language model (LLM) agentic performance. MAS strengthens task-specialized performance via role-based orchestration; RL leverages environment rewards to train stronger policies, such as Group Relative Policy Optimization (GRPO)-style optimization. Yet applying on-policy RL training to MAS is underexplored. While promising, 
it poses several challenges. On the algorithm side, Standard GRPO grouping assumptions fail in MAS because prompts differ by role and turn.
 On the system side, the training system 
needs to support 
MAS-workflow-based rollouts and on-policy updates for both single and multiple policy models.
To address these issues, 
we introduce \emph{\algname}, consisting of (i) an \underline{A}gent- and \underline{T}urn-wise grouped RL algorithm tailored for MAS and (ii) a system to support both single-policy and multi-policy training. Across game, plan, coding, and math tasks, 
\algname~ demonstrates substantial performance gains across diverse domains.  Especially on long-horizon planning tasks, AT-GRPO boosts accuracy from a 14.0--47.0\% single-agent RL baseline to 96.0--99.5\%. Furthermore, it improves reasoning performance, with an average gain of 3.87--7.62\% on coding and 9.0-17.93\% on math.  
\end{abstract}

\noindent
\faGithub\ \textbf{Code \& Environment:} \\
\url{https://github.com/pettingllms-ai/PettingLLMs}
\noindent


\section{Introduction}

Large Language Model (LLM) agents are task-specific workflows~\citep{Yao2023ReAct, xi2023rise,Wang2023LLMAgentsSurvey} that utilize LLMs as key components for decision making~\citep{shinn2023reflexion}, action taking~\citep{wang2023voyager}, and tool use~\citep{Qian2025ToolRL,schick2023toolformer}.
LLM agents 
have demonstrated strong promises across various application domains, such as embodied control \citep{ahn2022saycan,wang2023voyager},
software engineering \citep{tao2024magis,yu2025orcaloca}, 
expert drug discovery \citep{liu2024drugagent,inoue2024drugagent}, 
and scientific ideation and hypothesis testing \citep{ghafarollahi2024sciagents}.

Today, two complementary approaches are widely used to improve the performance of LLM agents:
multi-agent systems (MAS) and reinforcement learning (RL). RL treats the LLM as a policy and iteratively updates its weights to strengthen decision-making: at each iteration, the current model interacts with the environment, collects rule-based rewards, and then computes a policy optimization loss to update the parameters \citep{shao2024deepseekmath}. In practice, this workflow requires a training stack that supports both scalable rollouts and online updates, e.g., VERL \citep{sheng2025hybridflow} and AReaL \citep{fu2025areal}. 
MAS typically 
employs prompt-only augmentation on a shared LLM policy for role-based coordination; practical deployments instantiate diverse workflows. 
Recent studies~\citep{belcak2025slm_agentic,chen2024ioa,wang2024moa} further highlight the potential benefits of role-specialized MAS, which adopts distinct models for different roles, enabling role-specialized policies in inference. However, the effectiveness of RL training on role-specialized MAS is underexplored.

\begin{figure}[ht]
  \centering
  \includegraphics[width=\linewidth]{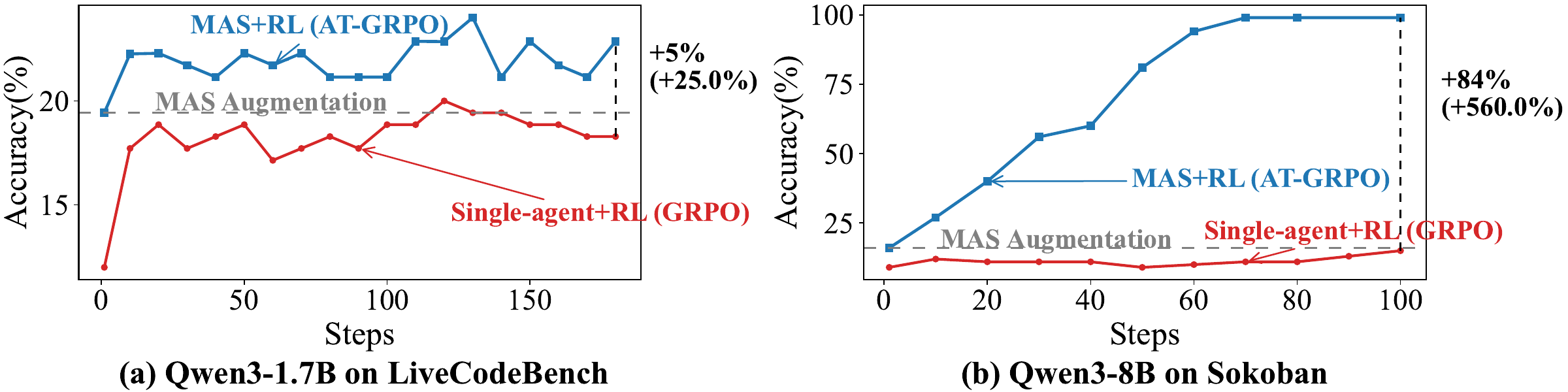}
  \vspace{-20pt}
  \caption{\textbf{MAS+\algname~vs.\ Single-agent+GRPO.} The gray line denotes the prompt-only MAS baseline.
  }
  \vspace{-20pt}
  \label{fig:myfig}
\end{figure}

A natural next step is to combine the two: using RL to train MAS, 
such that we gain both stronger learned policies, role-specialized collaboration. However, bringing RL into MAS raises two coupled challenges. 
First, training a MAS may require concurrently launching multiple models, orchestrating inter-agent environment interactions, and performing independent on-policy parameter updates. 
But most existing on-policy RL frameworks for LLM agents only support a single model \citep{volcengine2025verl,sheng2024hybridflow,fu2025areal}.
Second, rollouts from MAS are difficult to group. The advantage must be conditioned on interaction history and role to ensure fair credit assignment. Group-based RL objectives designed for a single agent \citep{volcengine2025verl,Qian2025ToolRL,feng2025group} are not directly applicable to MAS.

To address these challenges, we first design \emph{AT-GRPO}, an \underline{A}gent- and \underline{T}urn-wise grouped RL method that adapts group-relative optimization for MAS. Furthermore, we develop a novel training system 
to support on-policy RL for MAS. Our training system supports rollouts for diverse MAS workflows and enables on-policy RL training for both role-sharing policy and role-specific policies. 
We conduct extensive experiments on Qwen3 models across a range of representative agentic domains, including game, planning, coding, and mathematical reasoning. As highlighted in Fig.~\ref{fig:myfig}, \algname~ (blue) significantly outperforms single-agent GRPO (red). For instance, it achieves a 5.0\% higher accuracy (+25.0\% relative) on LiveCodeBench (with Qwen3-1.7B), while the improvement increases to 84.0\% on Sokoban (with Qwen3-8B). 

This paper makes the following key contributions:
\squishlist
    \item \textbf{AT-GRPO Algorithm.} We introduce an agent- and turn-wise grouped RL algorithm, AT-GRPO, and identify the substantial benefits of applying on-policy RL to MAS across diverse domains: planning, gaming, coding and mathematical reasoning tasks.
    \item \textbf{MAS Training System.} We design a novel training system to support  (i) executing rollouts for diverse MAS workflows and (ii) performing on-policy RL updates for multiple policies. 
    
    \item Our method delivers \textbf{consistent gains across diverse domains}. On long-horizon planning tasks, it overcomes a key bottleneck of single-agent RL, boosting accuracy from a 14--47\% baseline to 96.0-99.5\%. Furthermore, it also demonstrates gains on code and math benchmarks, with average improvements of 3.87--7.62\% and 9.0--17.93\%, respectively.
    \item 
    Our \textbf{analysis} shows that (1) RL training on MAS reinforces role-specific specialization; (2) with MAS \algname, whether to choose a role-sharing policy or role-specialized policies needs to 
    be determined by the task characteristics.
\squishend

\section{Related Work}

\textbf{Role-sharing vs. Role-specialized Policies in MAS.}
A predominant approach in LLM-based MAS centers on a role-sharing architecture, where a single policy is shared across all agents. In these frameworks, such as AutoGen~\citep{wu2023autogen} and MetaGPT~\citep{hong2024metagpt}, role-specific behavior is elicited at inference time via prompt augmentation. More recently, research has begun to explore role-specialized policies. This shift is motivated by the observation that a single LLM's performance exhibits significant variance across domains~\citep{chen2024ioa, wang2024moa, belcak2025slm_agentic}. Consequently, assigning distinct and more suitable models to specialized roles, as demonstrated by~\cite{ye2025xmas,belcak2025slm_agentic}, has emerged as a promising direction for enhancing performance. Despite this architectural evolution, recent surveys~\citep{pan2025whydomasfail, guo2024llmma} indicate that most studies focus on inference-time design, leaving the potential of training MAS policies with RL largely underexplored.


\textbf{RL Training for MAS.} 
\label{sec:relatedwork}
RL has become a key technique for LLMs agent training, using group-relative and rule-based rewards to enhance reasoning, long-horizon planning, game, and tool use \citep{feng2025group,wang2025ragen,Qian2025ToolRL,hu2025lmgame}. These approaches, however, predominantly operate within a single-agent framework.
While a growing body of work attempts to extend RL to Multi-Agent Systems (MAS), most efforts remain confined to limited interaction settings or fixed role structures. For instance,  CURE~\citep{wang2025cure} focuses on co-evolving a Coder and Unit-Tester using a role-sharing policy specifically for code generation. Similarly, SPIRAL~\citep{liu2025spiral} employs self-play in zero-sum games using a single LLM, while MHGPO~\citep{chen2025mhgpo} targets retrieval-augmented generation. MAPoRL~\citep{park2025maporl,park2025maporl2} and CoRY~\cite{ma2024cory} train LLMs within fixed, homogeneous-role debate workflows. More recent works also exhibit  limitations: \revised{MARFT~\cite{Liao2025MARFT} \footnote{We compare against MARFT v3, the latest preprint available prior to the completion of this work.}
restricts agents to single-turn sequential interactions, and MARTI~\cite{zhang2025marti} merely introduces basic single-agent RL algorithms (e.g., GRPO) to the MAS setting. Critically, the implemention for these works is concentrated on the single domain of math. We include a more comprehensive comparison of these related works in Appendix~\ref{app:mas-rl-comparison}. In contrast, our study offers a more comprehensive solution. We propose a general algorithm designed for multi-turn, multi-agent environments. Unlike prior works, we conduct a thorough analysis and evaluation of both shared and role-specific policies across diverse MAS workflows and varying domains.}

\section{Preliminaries}
\begin{figure}[t]     
  \centering
  \includegraphics[width=0.9\linewidth]{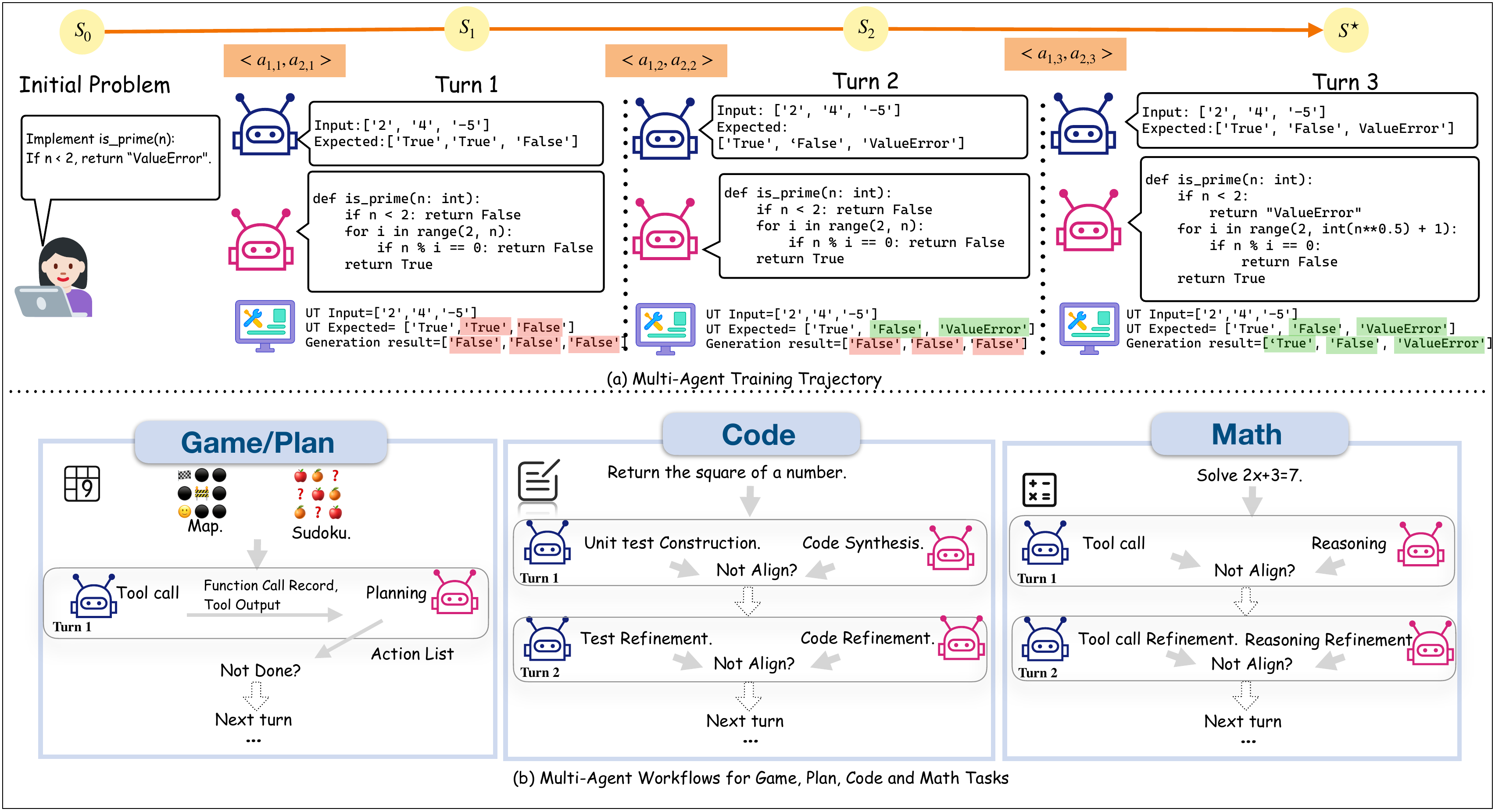} 
  \vspace{-10pt}
\caption{\textbf{MAS workflow across different domains.}
(a) Role-based coordination: code generation via a coder–tester loop.
(b) Different task-specific workflows for Game/Plan, Code, and Math; see Sec.~\ref{workflow in exp} and Appendix~\ref{app:prompt} for workflow details.}
\vspace{-10pt}

  \label{fig:mas-overview}
\end{figure}
\paragraph{\textbf{MAS Setting.}}
The $N$-agent LLM system is modeled as a Markov game $\mathcal{M}=(\mathcal{S},\{\mathcal{A}_i\}_{i=1}^N, \mathcal{T}, \{r_i\}_{i=1}^N, T, H)$, where $\mathcal{S}$ is the state space; $\mathcal{A}_i$ is the action space of agent $i$;  The transition function $\mathcal{T}$ induces intra-turn micro-transitions where $s_{t,0}=s_t$ and $s_{t,i} = \mathcal{T}(s_{t,i-1}, a_{t,i}, i)$, culminating in $s_{t+1}=s_{t,N}$.
The reward for agent $i$ is given by $r_i: \mathcal{A}_i \to [0, 1]$, and the turn horizon $T$, the optimization step horizon $H$. 
At each turn $t$, agent $i$ receives an observation summarizing the environment state and interaction history $h_t$,
$o_{t,i} \!=\! o_i(s_t, h_t)$.
Each agent $i$ is implemented with a role-specific prompt template $\mathsf{P}_i(\cdot)$.
Let $\Theta=\{\theta^{(m)}\}_{m=1}^M$ denote the set of LLM parameter vectors, with $1\le M\le N$, and let
$\sigma:\{1,\ldots,N\}\!\to\!\{1,\ldots,M\}$ assign each agent to an LLM.
We treat one LLM rollout (a token sequence) as a single macro-action $a_{t,i}$.
A \emph{turn} is one full interaction in which all agents emit macro-actions to the environment.
A \emph{step} denotes one optimization update to the parameter set $\Theta$ during training.
\paragraph{\textbf{MAS Workflow.}}
Following prior work \citep{wang2025cure,  ahn2022saycan, Chen2025CodeSteer}, we employ domain-specific MAS workflows, as shown in Fig.~\ref{fig:mas-overview}. Our experiments confirm that this prompt-only method outperforms a single-agent baseline (see Tab.~\ref{tab:qwen3-1p7b} and~\ref{tab:qwen3-8b} in Sec.~\ref{sec:exp}).
\paragraph{\textbf{Group-based RL.}}\label{sec:grpo}
Methods for LLM agentic training with group-relative advantages \citep{feng2025group,wang2025ragen,Qian2025ToolRL} operate by first sampling $K$ candidate actions $\{a_{t}^{\revised{(c)}}\}_{\revised{c=1}}^{K}$ for a given prompt. Each action is evaluated to obtain a rule-based reward $R(a_{t}^{\revised{(c)}})$, forming a comparison group:
$G= \big\{\, (a_{t}^{(1)}, R(a_{t}^{(1)})),\; \ldots,\; (a_{t}^{(K)}, R(a_{t}^{(K)})) \,\big\}.$ For each action $a_{t}^{\revised{(c)}}$ in this group, the relative advantage is then defined as its mean-centered and normalized return.
\begin{equation}
\label{Eq:GRPO_adv}
A_g\!\big(a_{t}^{\revised{(c)}}\big)
= \frac{\, R(a_{t}^{\revised{(c)}}) - \mathrm{mean}\!\left(\{\,R(a_{t}^{\revised{(c)}})\,\}_{\revised{c=1}}^{K}\right) \,}
     {\, F_{\mathrm{norm}}\!\left(\{\,R(a_{t}^{\revised{(c)}})\,\}_{\revised{c=1}}^{K}\right) \,},
\end{equation}

\paragraph{Role-sharing vs. Role-specialized Policy Optimization.}
We distinguish between two optimization regimes, role-sharing and role-specialized, both of which initialize policies from the same base model. During rollouts, each agent $i$ generates a dataset $\mathcal{D}_i$, which consists of sample groups. A single group $g$ is composed of a shared observation context $o_g$ and $K$ candidate actions with their corresponding advantages, denoted as $g=\{i, a_g^{(c)}, A_g^{(c)}\}_{c=1}^K$. The core difference between the two regimes lies in how the training data is batched. A minibatch $\mathcal{B}_m= \bigcup_{i \,:\, \sigma(i)=m} \mathcal{D}_i.$ for a specific policy $\theta^{(m)}$ is constructed by pooling the datasets from all agents assigned to it:

\begin{align}\mathcal{L}(\theta^{(m)}) \;=\;
    -\,\mathbb{E}_{g\in\mathcal{B}_m}\!\left[
        \frac{1}{K}\sum_{c=1}^K
        \min\!\Big(
            r_{g}^{(c,m)}(\theta^{(m)})\,A_g^{(c)}\,,
            \revised{\operatorname{clip}\!\big(r_{g}^{(c,m)}(\theta^{(m)}),\,1-\varepsilon,\,1+\varepsilon\big)\,A_g^{(c)}}
        \Big)
    \right]
    \label{eq:loss_func_grpo_clip}
\end{align}
where $r(\theta) = \frac{\pi_\theta(o_i|q)}{\pi_{\theta_{old}}(o_i|q)}$.
\noindent \emph{Role-sharing policy ($M{=}1$):} All agents share a single policy $\theta^1$. The training batch is the union of data from all agents, $\mathcal{B}_1=\bigcup_{i=1}^N \mathcal{D}_i$, and is used for a single joint update: $\theta^1 \leftarrow \theta^1 - \eta\nabla_{\theta^1} \mathcal{L}(\theta^1)$.

\noindent \emph{Role-specialized policies ($M=N$):} Each agent $i$ has a distinct policy $\theta^{(i)}$, such that $\sigma(i)=i$. Each policy is updated independently on $\mathcal{B}_i = \mathcal{D}_i$, and update policy: $
\theta^{(i)} \leftarrow \theta^{(i)} - \eta \nabla_{\theta^{(i)}} \mathcal{L}(\theta^{(i)})$.

\section{Method}

\subsection{Algorithm Design: \algname}

\begin{wrapfigure}{t}{0.65\textwidth} 

  \centering
  
   \vspace{-20 pt}
  \includegraphics[width=0.65\textwidth]{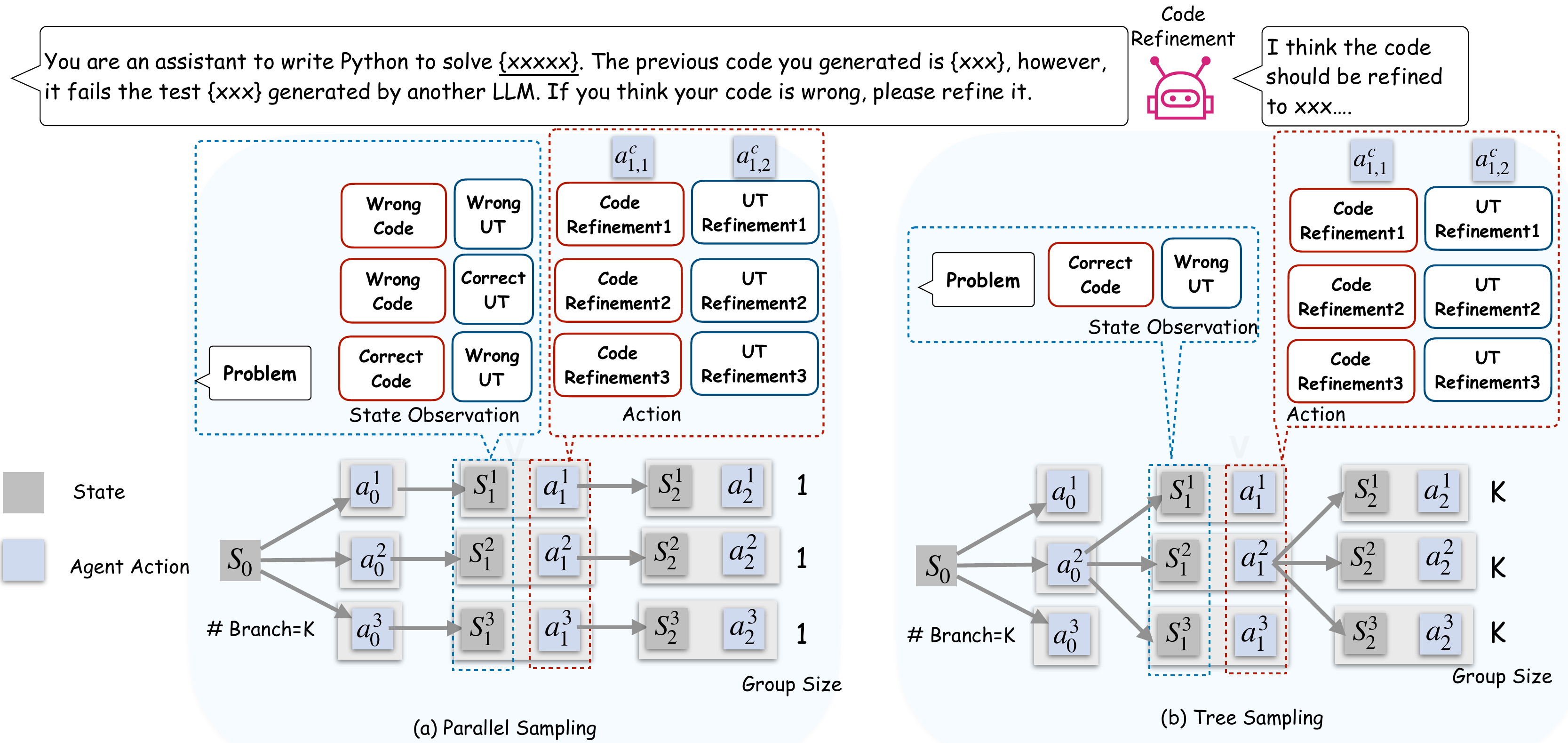}
   \vspace{-20 pt}
  
  \caption{
  \label{fig:group_explain}
\revised{Two sampling schemes.}
\textbf{(a) In parallel sampling},  trajectories are sampled but incomparable, leading to groups of size 1.
\textbf{(b) In tree sampling}, branching at each turn forms a valid comparison group of size $K$.}

\vspace{-10 pt}
\end{wrapfigure}

GRPO's advantage calculation (Eq.~\ref{Eq:GRPO_adv}) hinges on a fair comparison among all candidates within a group. This fairness is enforced by the reward mechanism itself. As illustrated in Fig.~\ref{fig:mas-overview} (top), token-level scoring assigns credit to the generated response tokens (Reward Mask=1), while the prompt tokens receive no credit (Reward Mask=0). Since the advantage is determined solely by the quality of the response, a valid and fair comparison is only possible when all responses in a group originate from an identical prompt. Consequently, single-agent LLM-RL methods\citep{wang2025ragen,Qian2025ToolRL,feng2025group} typically form groups by sampling multiple responses to the same question.

In MAS, however, a ``prompt'' is not only a question description, but also embeds the role-specific context and cross-agent interaction history. For example, \revised{in the code tasks depicted in Fig.~\ref{fig:mas-overview}~(a) , where the workflow entails a coder-tester loop: one agent synthesizing code and the other creating unit tests, and they iteratively refine the output until alignment} (Fig.~\ref{fig:group_explain}, \revised{top}), the turn-2 refinement prompt already contains the turn-1 code, unit tests, and role-specific prompt format, so prompts differ across turns and roles. We therefore adopt agent-wise and turn-wise grouping as a natural extension of tabular-wise group-normalized advantages in GiGPO~\citep{feng2025group} to the multi-agent setting: candidates share the same role and turn position,  ensuring prompt identity for valid GRPO advantage comparisons.

However, agent- and turn-wise grouping introduces a new question. If we follow the common parallel sampling used by prior agentic RL—sample $K$ full trajectories from the initial state/problem (Fig.~\ref{fig:group_explain}~(a), bottom), each group size $=1$ when $t>1$: no other sample shares the identical prompt. GRPO therefore eliminates its variance-reduction effect and yields unstable updates. To address these challenges, we develop \algname\, (see Alg.~\ref{alg:hatgrpo_mas}) with three key ideas: \emph{tree-structured sampling}, \emph{agent– and turn-wise grouping}, and \emph{agent-wise credit assignment}.

\begin{algorithm}[t]
\caption{\algname: Agent- and Turn-wise MAS RL Training}
\label{alg:hatgrpo_mas}
\begin{algorithmic}[1]
\Require Markov game $\mathcal{M}$, policies $\Theta=\{\theta^{(m)}\}_{m=1}^M$, role mapping $\sigma$, sampling temperature $T_{\text{samp}}$, branches $K$, total steps $S$, batch size $E$, turn horizon $T$, termination condition $\mathcal{I}_{\mathrm{term}}$.
\Statex\textit{/*-- Termination helper: returns true if horizon reached or env signals done --*/}

\For{training step $s = 1, \dots, S$}
  \Statex\textit{/*-- Phase 1: On-Policy Rollout \& Data Collection --*/}
  \State Initialize per-agent datasets $\{\mathcal{D}_i\}_{i=1}^N \gets \emptyset$. Resample $E$ environments.
  \For{\textbf{each} environment instance $e \in \{1, \dots, E\}$ \textbf{in parallel}}
    \For{$t = 0$ \textbf{to} $T-1$}
      \State $s_{t,0,e} \gets s_{t,e}$ \Comment{Initialize micro-step state}
      \For{\textbf{each} agent $i \in \{1, \dots, N\}$}
        \State $\forall c\in \{1,\dots,K\}$, $a^{(c)}_{t,i,e} \sim \pi_{\theta^{(\sigma(i))}}(\cdot \mid o_{t,i,e};\, T_{\text{samp}})$; compute $r^{(c)}_{t,i,e}$ (Eq.~\ref{eq:reward})
        \State Define group key $g \gets \text{hash}(e, i, t)$ and compute advantages $\{A^{(c)}_g\}_{c=1}^K$ (Eq.~\ref{Eq:GRPO_adv}). 
        \State Append $(g, o_{t,i,e}, \{a^{(c)}_{t,i,e}\}_{c=1}^K, \{A^{(c)}_g\}_{c=1}^K)$ to $\mathcal{D}_i$.
        \State $c^\star \gets \arg\max_{c} r^{(c)}_{t,i,e}$;\quad $a_{t,i,e} \gets a^{(c^\star)}_{t,i,e}$. {\color{darkblue}\textit{(Tree-structured sampling.)}}
        \State $s_{t,i,e} \gets \mathcal{T}\big(s_{t,i-1,e}, a_{t,i,e}, i\big)$ \Comment{Agent-wise micro-transition}
      \EndFor
      \State $s_{t+1,e} \gets s_{t,N,e}$ \Comment{End-of-turn state}
      \If{$\mathcal{I}_{\mathrm{term}}(s_{t+1,e})$} \textbf{break} \EndIf
    \EndFor
  \EndFor

  \Statex\textit{/*-- Phase 2: Per-Model Policy Update --*/}
  \For{\textbf{each} model $m \in \{1, \dots, M\}$ \textbf{in parallel}}
    \State Construct per-model batch $\mathcal{B}_m$, loss $\mathcal{L}(\theta^{(m)})$ on $\mathcal{B}_m$ using Eq.~\ref{eq:loss_func_grpo_clip} and update policy $m$.
  \EndFor
\EndFor
\end{algorithmic}
\end{algorithm}

\paragraph{Tree-structured Sampling.}
At each turn~$t$, for each agent~$i$, we sample~$K$ candidate actions and their corresponding rewards from the current state (Alg.~\ref{alg:hatgrpo_mas}, line~7). The advantages for these~$K$ candidates are then calculated within this group (line~9). Subsequently, the full data tuple---containing the group key, observation, $K$~actions, and their~$K$~advantages---is added to a dataset~$D_i$ specific to the policy of the acting agent $i$ (line~10). To proceed with the environment rollout, we greedily select the candidate with the highest reward to be the executed action (line~11). This greedy selection strategy concentrates exploration on coordination-critical decisions and helps maintain a balanced mix of positive and negative samples, which stabilizes the learning optimization.

\paragraph{Agent– and Turn-wise Grouping.}
We group experiences based on the acting agent and the turn number within each parallel environment instance. Operationally, we implement this by defining a unique group key~$g$ for each agent~$i$ at each turn~$t$ in each environment~$e$ using a lightweight hash function (Alg.~\ref{alg:hatgrpo_mas}, line~8). All data generated from the $K$-branch sampling at that step, including the observation and the calculated advantages, is stored together under this group key (line~10). During the policy update phase, these collected data groups are used to construct per-model training batches for the final optimization step (lines~20--21).
\paragraph{Agent-wise Credit Assignment.}
Inspired by mixed-reward designs in cooperative Multi-Agent RL \citep{Mao2020RewardDesignMARL,Sheikh2020DEMADDPG}, we assign credit using a mixture of global and local rewards. At each turn~$t$, the environment provides a global team reward~$r^{\mathrm{team}}$ and an agent-specific local reward~$r^{\mathrm{loc}}_{i}$ that evaluates its subtask performance. These components are combined using a hyperparameter~$\alpha$ to form the final reward for agent~$i$:
\begin{equation}
\label{eq:reward}
    r_{t,i} \;=\; \alpha\, r^{\mathrm{team}}_t \;+\;  r^{\mathrm{loc}}_{t,i}
\end{equation}

This formulation balances a shared team objective with role-specific incentives. For instance, in a coder-tester MAS, the team reward~$r^{\mathrm{team}}$ is the pass rate of the generated program on a set of golden unit tests. The local rewards~$r^{\mathrm{loc}}_i$ are tailored to each role: the coder is rewarded for its own code's pass rate, while the tester is rewarded based on the pass rate of a golden reference implementation against its generated tests. Detailed reward designs for all tasks are provided in Appendix~\ref{app:training_detail}.

\subsection{MAS Training System}

\begin{wrapfigure}{r}{0.5\textwidth}
  \vspace{-40pt}
  \centering
  \includegraphics[width=\linewidth]{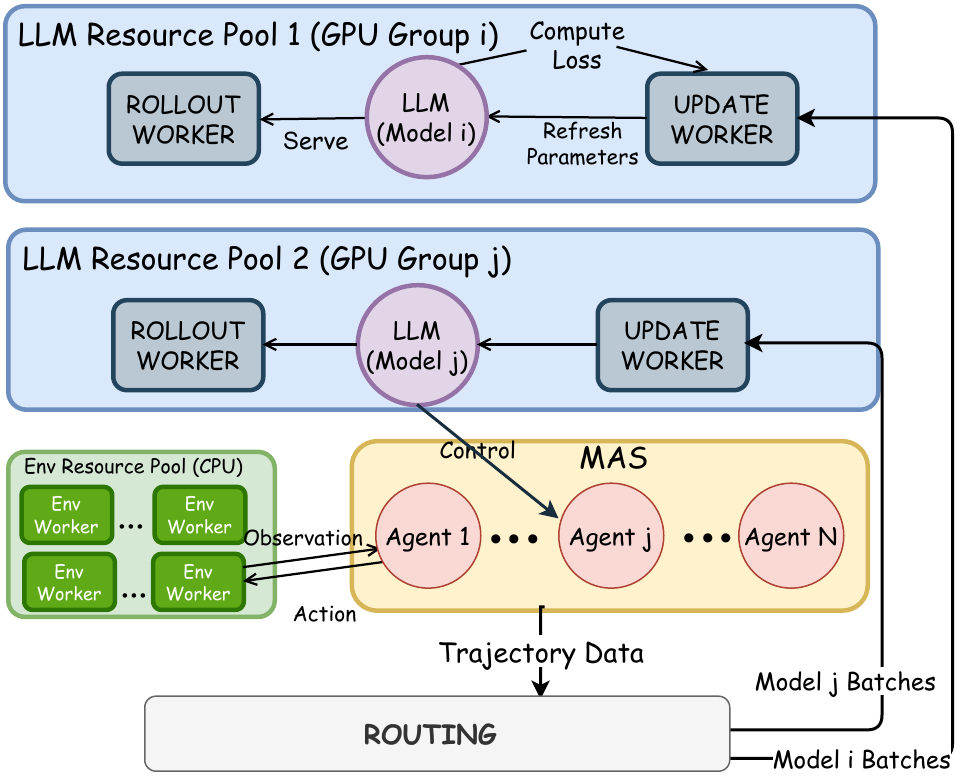}
  \vspace{-18pt}
    \caption{\textbf{ MAS training system.}
  Each LLM $m$ has a GPU-pinned model pool with a RolloutWorker and an UpdateWorker. A CPU environment pool hosts envworkers that execute environment steps. Trajectories are routed to the corresponding UpdateWorker.}
  \label{fig:mas-sys}
  \vspace{-28pt}
\end{wrapfigure}
Mainstream RL post-training frameworks for LLMs, e.g., TRL~\citep{vonwerra2020trl}, VERL~\citep{sheng2024hybridflow}, AReaL~\citep{fu2025areal}, and OpenRLHF~\citep{hu2024openrlhf} primarily support single-agent RL training, which typically involves: a single agent-environment interaction pattern, a single policy operating on a single data buffer, and a single LLM resource pool. This makes it difficult to (i) train multiple models in on-policy RL, (ii) maintain clean on-policy training data, and (iii) support diverse MAS workflow. 

We introduce a novel MAS training system to overcome these challenges and enable \algname~. By allocating an independent resource pool to each model, our system is designed to support the concurrent on-policy training of multiple policies. The system, depicted in Fig.~\ref{fig:mas-sys}, consists of the following components:

\paragraph{LLM Resource Pools (GPU).}
Each policy is managed within an independent resource pool. Following HybridFlow-style ~\citep{sheng2025hybridflow}, each pool comprises two workers: a \emph{RolloutWorker} for inference and an \emph{UpdateWorker} for optimization. During the rollout phase, all policies interact collectively according to the Alg.~\ref{alg:hatgrpo_mas} and MAS workflow; Once collected, each trajectory is routed to the corresponding UpdateWorker, maintaining an on-policy learning regime for every policy.

\paragraph{Environment Execution (CPU) and Data Flow.}
Environment steps run in a fleet of CPU \emph{EnvWorkers}, each managing a single sandboxed instance to ensure safety and reproducibility (seeding, wall-clock timeouts, IO quotas, and deterministic tool harnesses). This one-actor-per-instance mapping efficiently supports thousands of concurrent rollouts in parallel. EnvWorkers stream observations, tool logs, and rule-based rewards to a \emph{Router}. The Router dispatches collected experience based on policy assignment: experiences generated by an agent $i$ are sent to the Updateworker of its designated policy $\sigma(i)$.  

\section{Experiments}

\subsection{Datasets and models.}
\label{sec:exp-setting}
1. \textbf{Experimental Setup.} 
We train and evaluate Qwen3 models at 1.7B and 8B in the no-thinking mode~\citep{Qwen3TR}. All runs use a single node with \(8\times\)~H100 GPUs.  The rollout sample size is \(K{=}4\) and the turn horizon is $T=4$ \revised{for both multi-agent (MA) and single-agent (SA) settings}. The reward-mixing coefficient is \(\alpha{=}1\) without further tuning. Full training details appear in Appendix~\ref{app:training_detail}.

\paragraph{2.\ Baselines.} We evaluate five variants (all initialized from the same base model): (a) Single Agent (prompt-only): one frozen LLM solves the task end-to-end; (b) Single Agent + GRPO: as (a) but trained with GRPO~\citep{shao2024deepseekmath}; (c) MAS (prompt-only): role-specialized prompting over a frozen, role-sharing backbone; (d) MAS + RL (role-sharing policy): all roles share one policy and pooled trajectories update it jointly; (e) MAS + RL (role-specialized policies): samples are routed by role and each policy is optimized independently (no parameter sharing).

\paragraph{3.\ Task Setups and Baselines.}
\label{workflow in exp}
\revised{To ensure a fair comparison, we align all environmental observations and reward signals across both MA and SA settings. While both paradigms utilize the same role-specific reward functions, the sole distinction is that the MA framework involves multiple agents capable of discussion.} Detailed prompt templates and reward specifications are provided in Appendix~\ref{app:reward} and~\ref{app:prompt}.

\revisedbegin

\noindent \textbf{Single Agent Variants.} We also evaluated a multi-turn SA variant for both Code and Math (see Appendix~\ref{sec:samt}), where a single agent repeatedly revises its own output until self-consistency. This setup is inherently less natural: the agent receives no additional environmental signal or cross-agent feedback, and the interaction pattern deviates from the QA-style pretraining regime of LLMs. Reflecting this mismatch, the multi-turn SA variant brought no empirical improvement and sometimes slightly degraded performance relative to the standard single-turn SA baseline.

\noindent \textbf{Code.} The environment observation is restricted to the problem description. The MA setting employs a dual-role debating mechanism: a \textit{Tester} generates unit tests and a \textit{Coder} generates the code. They refine their outputs iteratively until alignment is reached or the maximum turn limit is met. The natural SA baseline employs a \textit{Coder} to generate the solution directly, as no other environmental feedback and other agent's output is available.

\noindent \textbf{Math.} The environment observation consists is restricted to the problem.  MA uses Dual-role debating MAS: a \textit{Tool-User} (utilizing code interpreters) and a \textit{Reasoner} (performing direct reasoning) until alignment or the maximum turn limit is met. The SA setting uses a  \textit{Reasoner} to derive the answer directly, as no other environmental feedback and other agents' response.

\noindent \textbf{Planning and Game.} The environment observations are game states of the current turn. The MA setting employs a collaboration mechanism: a \textit{Tool-User} (executing the tools) and an \textit{Executor} (verifying tool outputs and executing actions). The SA employs an \textit{Executor} to perform actions. Both settings share identical termination conditions based on goal satisfaction or the turn budget.
\revisedend

\paragraph{4. Training and Evaluation Datasets.}

\noindent \textbf{Sudoku and Sokoban.} We evaluate our method on gaming tasks: a \(4{\times}4\) Sudoku and a \(6{\times}6\) Sokoban. We use instances with an automatic checker, following the symbolic task setup of \textsc{SymBench}~\citep{Chen2025CodeSteer}. To ensure a fair evaluation, we generate distinct training and validation sets using different random seeds and verify there is no overlap.

\noindent \textbf{Plan-Path.}
We use a \(10{\times}10\) grid-based Plan-Path environment. This follows the checker-backed symbolic task setup in CodeSteer's SymBench \citep{Chen2025CodeSteer}. To separate training and validation, we generate the two splits with distinct random seeds and verify no duplication.

\noindent\textbf{Code Generation.}
For training, we adopt size-specific corpora: the 1.7B Qwen model is trained on the APPS training split (introductory-difficulty subset) \citep{Hendrycks2021APPS}, while the 8B model is trained on CodeContests \citep{DeepMind2024CodeContests}. For model-generated code, we use the dataset’s golden unit tests to score correctness; for model-generated UT, we use the dataset’s golden reference solutions to compute the reward.
For evaluation, we use three widely adopted coding benchmarks spanning interview-style and contest-style settings:
APPS~\citep{Hendrycks2021APPS},
LiveCodeBench-v6~\citep{white2024livebench},
and CodeContests~\citep{DeepMind2024CodeContests}.

\noindent\textbf{Mathematical Reasoning.}
We train on the Polaris-Dataset-53K~\citep{Polaris2025} and evaluate on several standard mathematical reasoning benchmarks. For validation, we use \textsc{AIME24}/\textsc{AIME25} \citep{AIME2024,AIME2025} and \textsc{OlympiadBench} \citep{He2024OlympiadBench}. All math tasks use verifier-checked numeric scoring.

\subsection{Results and Analysis}
\label{sec:exp}
\begin{table}[t]
\centering


\resizebox{1\linewidth}{!}{%

\scriptsize
\setlength{\tabcolsep}{3.2pt}
\renewcommand{\arraystretch}{1.05}

\begin{threeparttable}
\caption{\textbf{Qwen3 1.7B} results on game, planning, coding, and math.}
\label{tab:qwen3-1p7b}
\begin{tabular}{ L{2.2cm} T{1.0cm} T{1.0cm} | T{1.2cm} | T{1.5cm} T{1cm} T{1.5cm} | T{1cm} T{1cm} T{1.5cm} }
\toprule
& \multicolumn{2}{c|}{\textbf{Game}} & \multicolumn{1}{c|}{\textbf{Plan}} & \multicolumn{3}{c|}{\textbf{Code}} & \multicolumn{3}{c}{\textbf{Math}} \\
\textbf{Method} & \textbf{Sudoku} & \textbf{Sokoban} & \textbf{Plan-Path} & \textbf{LiveCodeBench} & \textbf{APPS} & \textbf{CodeContests} & \textbf{AIME24} & \textbf{AIME25} & \textbf{OlympiadBench} \\
\midrule
\makecell[tl]{Single agent} & \sd{7.00}{0.00} & \sd{0.00}{0.00} & \sd{5.00}{0.00} & \sd{11.60}{0.00} & \sd{16.20}{0.00} & \sd{3.60}{0.00} & \sd{13.40}{0.00} & \sd{9.80}{0.00} & \sd{22.20}{0.00} \\
\makecell[tl]{Single agent + GRPO} & \sd{29.00}{22.00} & \sd{3.00}{3.00} & \sd{11.00}{6.00} & \second{\sd{18.80}{7.20}} & \sd{17.00}{0.80} & \sdneg{3.00}{-0.60} & \sdneg{10.00}{-3.40} & \sdneg{6.70}{-3.10} & \sd{23.80}{1.60} \\
\makecell[tl]{MAS} & \sd{69.00}{62.00} & \sd{0.00}{0.00} & \sd{10.00}{5.00} & \sd{19.00}{7.40} & \sd{16.60}{0.40} & \sd{3.60}{0.00} & \second{\sd{13.30}{-0.10}} & \sd{13.00}{3.20} & \second{\sd{35.90}{13.70}} \\
\revised{\makecell[tl]{MAS + GRPO}} 
  & \revised{\sd{87.00}{80.00}} 
  & \revised{\sd{1.00}{1.00}} 
  & \revised{\sd{82.00}{77.00}} 
  & \revised{\sd{20.60}{9.00}} 
  & \revised{\sd{17.60}{1.40}} 
  & \revised{\sd{4.80}{1.20}} 
  & \revised{\sd{13.30}{-0.10}} 
  & \revised{\sd{16.70}{6.90}} 
  & \revised{\sd{35.00}{12.80}} \\

\midrule
\makecell[tl]{MAS + \algname{}\\ w/ shared policy} & \best{\sd{99.00}{92.00}} & \second{\sd{10.00}{10.00}} & \second{\sd{96.00}{91.00}} & \sd{20.90}{9.30} & \second{\sd{17.60}{1.40}} & \second{\sd{4.80}{1.20}} & \best{\sd{16.70}{3.30}} & \second{\sd{16.70}{6.90}} & \best{\sd{39.60}{16.80}} \\
\makecell[tl]{MAS + \algname{}\\ w/ per-role policies} & \best{\sd{99.00}{92.00}} & \best{\sd{11.50}{11.50}} & \best{\sd{97.00}{92.00}} & \best{\sd{24.00}{12.40}} & \best{\sd{18.60}{2.40}} & \best{\sd{7.80}{4.20}} & \second{\sd{13.30}{-0.10}} & \best{\sd{18.30}{8.50}} & \sd{35.20}{13.00} \\
\bottomrule
\end{tabular}

\end{threeparttable}

}

\vspace{6pt}

\resizebox{\linewidth}{!}{%

\scriptsize
\setlength{\tabcolsep}{3.2pt}
\renewcommand{\arraystretch}{1.05}
\begin{threeparttable}
\caption{\textbf{Qwen3 8B} results on game, planning, coding, and math.}
\label{tab:qwen3-8b}
\begin{tabular}{ L{2.2cm} T{1.0cm} T{1.0cm} | T{1.2cm} | T{1.5cm} T{1cm} T{1.5cm} | T{1cm} T{1cm} T{1.5cm} }
\toprule
& \multicolumn{2}{c|}{\textbf{Game}} & \multicolumn{1}{c|}{\textbf{Plan}} & \multicolumn{3}{c|}{\textbf{Code}} & \multicolumn{3}{c}{\textbf{Math}} \\
\textbf{Method} & \textbf{Sudoku} & \textbf{Sokoban} & \textbf{Plan-Path} & \textbf{LiveCodeBench} & \textbf{APPS} & \textbf{CodeContests} & \textbf{AIME24} & \textbf{AIME25} & \textbf{OlympiadBench} \\
\midrule
\makecell[tl]{Single agent} & \sd{48.00}{0.00} & \sd{9.00}{0.00} & \sd{12.00}{0.00} & \sd{22.80}{0.00} & \sd{30.20}{0.00} & \sd{15.75}{0.00} & \sd{18.30}{0.00} & \sd{20.00}{0.00} & \sd{55.00}{0.00} \\
\makecell[tl]{Single agent + GRPO} & \sd{54.00}{6.00} & \sd{14.00}{5.00} & \sd{47.00}{35.00} & \sd{25.70}{2.90} & \sd{37.00}{6.80} & \sdneg{12.12}{-3.63} & \sd{18.30}{0.00} & \sd{26.67}{6.67} & \sdneg{54.80}{-0.20} \\
\makecell[tl]{MAS} & \sd{72.00}{24.00} & \sd{16.00}{7.00} & \sd{71.00}{59.00} & \sd{28.00}{5.20} & \sd{44.40}{14.20} & \sd{17.60}{1.85} & \sd{36.60}{18.30} & \sd{30.00}{10.00} & \sd{56.50}{1.50} \\
\revised{\makecell[tl]{MAS + GRPO}}
  & \revised{\sd{99.00}{51.00}}
  & \revised{\sd{30.00}{21.00}}
  & \revised{\sd{96.00}{84.00}}
  & \revised{\sd{24.20}{1.40}}
  & \revised{\sd{40.20}{10.00}}
  & \revised{\sdneg{10.30}{-5.45}}
  & \revised{\sd{33.30}{15.00}}
  & \revised{\sd{26.67}{6.67}}
  & \revised{\sdneg{53.20}{-1.80}} \\

\midrule
\makecell[tl]{MAS + \algname{}\\ w/ shared policy} & \best{\sd{99.50}{51.50}} & \second{\sd{96.00}{87.00}} & \second{\sd{93.00}{81.00}} & \second{\sd{30.28}{7.48}} & \second{\sd{45.80}{15.60}} & \best{\sd{18.10}{2.35}} & \second{\sd{50.00}{31.70}} & \second{\sd{35.20}{15.00}} & \best{\sd{56.80}{1.80}} \\
\makecell[tl]{MAS + \algname{}\\ w/ per-role policies} & \second{\sd{99.00}{51.00}} & \best{\sd{98.00}{89.00}} & \best{\sd{96.00}{84.00}} & \best{\sd{33.10}{10.30}} & \best{\sd{46.50}{16.30}} & \best{\sd{18.10}{2.35}} & \best{\sd{57.00}{38.70}} & \best{\sd{40.00}{20.00}} & \second{\sd{56.60}{1.60}} \\
\bottomrule
\end{tabular}
\begin{tablenotes}
\footnotesize
\item Parentheses denote gain over the Single Agent baseline; \colorbox{gold!25}{best} and \colorbox{silver!25}{second-best} results per column are highlighted.
\end{tablenotes}
\end{threeparttable}

}
\vspace{-10pt}
\end{table}

We evaluate AT-GRPO across four distinct domains (game, planning, code, and math) using two model scales (Qwen3 1.7B and 8B). To contextualize its performance, we benchmark against all the variants described in Sec.~\ref {sec:exp-setting}. Tab.~\ref{tab:qwen3-1p7b} and Tab.~\ref{tab:qwen3-8b} summarize our main results.

\paragraph{\textbf{MAS + \algname{} consistently yields substantial performance gains, especially in long-horizon planning tasks.}}
 MAS + \algname{} elevates the success rate from a 14--47\% range for the single-agent baseline to 96.0--99.5\%. By analyzing the dialogue records between agents, we find this dramatic improvement stems from an emergent collaboration: the tool agent learns to generate correct algorithms (e.g., BFS, $A^{\star}$ search), while the plan agent provides crucial oversight, interpreting execution outcomes and delivering the corrective final action list.  On-policy RL training within the MAS enhances inter-agent coordination. Conversely, training agents in isolation results in only limited improvement, as detailed in our ablation study (Sec.~\ref{sec:ablation-study}, Tab.~\ref{tab:pp-ablate}). 
Furthermore, on the coding and math benchmarks, our approach yields consistent gains, with absolute gains over the baseline ranging from +2.35 (CodeContests) to +16.30 (APPS) in coding, and from +1.80 (OlympaidBench) to +38.70 (AIME24) in math. We hypothesize two reasons: (1) Base models like Qwen3 have already been extensively trained for these common domains, as noted in their official reports~\citep{Qwen3TR}, potentially leading to performance saturation. (2) The diverse nature of problems within these domains presents a greater challenge for improvement via RL training.

\textbf{With MAS \algname, whether choosing a role-sharing policy or role-specialized policies should
be determined by the task characteristics.}
Role-specialized policies involve a fundamental trade-off: training each agent exclusively on its own data fosters deep specialization, but prevents access to potentially useful data from other roles. Our findings indicate that the optimal resolution to this trade-off depends on the task characteristics. We observe clear benefits for role specialization in the coding domain, where the Coder and Tester functions are highly distinct. This separation allows each agent to hone its specific skills, improving the average accuracy by 3.05 points with the Qwen3 1.7B.In contrast, the roles in the math domain exhibit greater functional overlap, meaning a shared policy can sometimes be superior. For instance, with the Qwen3 1.7B model on OlympiadBench, the shared policy achieves a 39.60\% accuracy, surpassing the 35.20\% from per-role policies. This suggests the Tool agent, which must often perform reasoning to execute tool calls, benefits from the Reasoner's training data.  For game/plan tasks, this choice becomes moot, as both configurations already achieve near-optimal, saturated performance (e.g., 99.50 on Sudoku).
\revisedbegin
\begin{wrapfigure}{r}{0.5\textwidth} 
    \vspace{-0.8em} 
    \centering
    \includegraphics[width=0.48\textwidth]{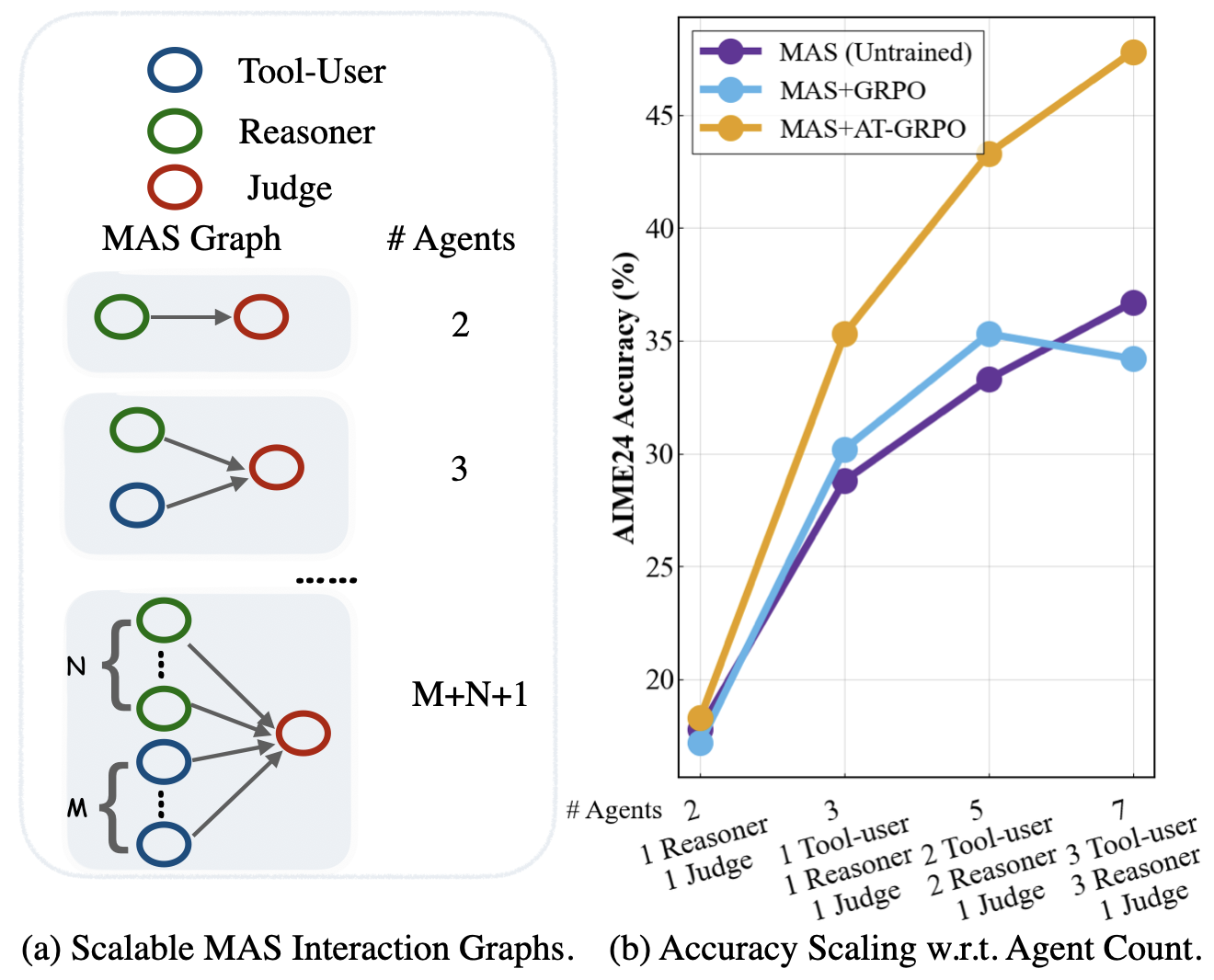}
    \vspace{-0.6em}
    \caption{
        (a) The system aggregates outputs from an ensemble of $N$ Reasoners and $M$ Tool-Users into a Judge. The total agent count scales as $M+N+1$, allowing for flexible resource allocation. 
        (b) Evaluation on AIME24 (using Qwen3-8B).
    }
    \label{fig:mas_scaling}
    \vspace{-10pt}
\end{wrapfigure}

\vspace{-20pt}
\paragraph{{Limitations of MAS-GRPO.}} 
Empirical results in Tab.~\ref{tab:qwen3-1p7b} and~\ref{tab:qwen3-8b} indicate that directly applying GRPO to MAS often results in performance degradation. Notably, Qwen3-8B exhibits suboptimal results CodeContests (17.60 $\to$ 10.30) and OlympiadBench (56.50 $\to$ 53.20). We attribute this to the violation of the identical-state assumption: as multi-turn interaction histories diverge, the group-averaged baseline incorrectly aggregates heterogeneous states. This structural misalignment biases advantage estimation and destabilizes optimization.

\textbf{Scalability Analysis with Collaborative Agents.}
To investigate scalability, we deploy a modular MAS architecture comprising Reasoners, Tool-Users, and a Judge (Fig.~\ref{fig:mas_scaling}(a)). By varying the number of concurrent Reasoners ($N$) and Tool-Users ($M$), we scale the total agent count ($M+N+1$) to expand the exploration space. Fig.~\ref{fig:mas_scaling}(b) demonstrates the scalability of MAS+AT-GRPO. While the baseline MAS+GRPO fails to scale effectively—saturating at 34.1\% accuracy in the 7-agent regime—our method successfully leverages the increased ensemble size, achieving a continuous performance gain from 18.2\% to 47.7\%. This confirms that MAS+AT-GRPO can effectively scale across multiple agents without hitting the coordination bottlenecks observed in baselines. For a detailed analysis of computational efficiency and complexity, please refer to Appendix~\ref{app:complexity}.

\begin{wraptable}{r}{0.55\textwidth}
    \vspace{-12pt}
    \centering
    \caption{Comparison with existing MARL frameworks. 
    We report Accuracy (\%) for math/logic tasks and Pass@1 (\%) for code tasks.}
    \label{tab:marl_comparison_all}
    \setlength{\tabcolsep}{4pt}
    \scriptsize
    \resizebox{\linewidth}{!}{%
    \begin{tabular}{l l l c}
    \toprule
    \multicolumn{4}{c}{\textbf{Math: Accuracy (\%)}} \\
    \midrule
    \textbf{Backbone} & \textbf{Method} & \textbf{Config} & \textbf{Acc.} \\
    \midrule
    Phi-3-mini 
      & Vanilla Baseline & Zero-shot & 65.0$^*$ \\
      & MAPORL           & Trained   & 81.0$^*$ \\
      & \textbf{Ours (MAS)} 
                        & \textbf{Untrained} 
                                     & \textbf{84.4} \\
      & \textbf{Ours (MAS+AT-GRPO)} 
                        & \textbf{Trained} 
                                     & \textbf{88.7} \\
    \midrule
    Qwen2.5-Coder-3B-Instruct
      & Vanilla Baseline & Zero-shot & 76.8$^*$ \\
      & MARFT            & Trained   & 78.7$^*$ \\
      & \textbf{Ours (MAS)} 
                        & \textbf{Untrained} 
                                     & \textbf{84.4} \\
      & \textbf{Ours (MAS+AT-GRPO)} 
                        & \textbf{Trained} 
                                     & \textbf{87.1} \\
    \midrule
    \multicolumn{4}{c}{\textbf{Code: Pass@1 (\%)}} \\
    \midrule
    \textbf{Backbone} & \textbf{Method} 
                      & \multicolumn{1}{c}{\textbf{CodeContests}} 
                      & \textbf{LiveCodeBench} \\
    \midrule
    Qwen-2.5-7B-Instruct 
      & Vanilla Baseline & 22.8$^*$ & 26.9$^*$ \\
      & CURE             & 25.9$^*$ & 31.2$^*$ \\
      & \textbf{Ours (MAS)} 
                         & \textbf{30.3} & \textbf{30.4} \\
      & \textbf{Ours (+AT-GRPO)} 
                         & \textbf{34.2} & \textbf{35.3} \\
    \bottomrule
    \multicolumn{4}{r}{\tiny $^*$ Results cited from original papers.} \\
    \end{tabular}
    }
    \vspace{-16pt}
\end{wraptable}

\subsection{Comparison with other MARL Frameworks}

To assess the efficacy of our framework, we conducted ablation studies against representative baselines: As summarized in Tab.~\ref{tab:marl_comparison_all}. we benchmark against three representative baselines: MAPORL~\citep{park2025maporl2}, MARFT~\citep{Liao2025MARFT}, and CURE~\cite{wang2025cure}. For fair comparison, we utilize identical base models and dataset splits. Our analysis highlights the advantages of two distinct MAS features: \textbf{heterogeneous agent roles} and \textbf{multi-turn iterative interaction}, as summarized in Tab.~\ref{tab:marl_comparison_all}.

\textbf{Comparison with MAPORL.}
 We compare our approach with MAPORL using the Phi-3-mini-128k (3.4B) on gsm8k dataset~\cite{cobbe2021training}. While MAPORL relies on a debating mechanism among homogeneous agents, our framework realizes role heterogeneity—synergizing a \textit{Reasoning Agent} with a \textit{Tool-use Agent} for verification. This structural specialization proves superior: our untrained MAS achieves 84.4\%, outperforming the trained MAPORL (81.0\%). With AT-GRPO training, our performance further improves to 88.7\%.

\textbf{Comparison with MARFT and CURE.} 
This comparison highlights the critical efficacy of iterative alignment over single-turn workflows. In math reasoning ( Qwen2.5-Coder-3B-Instruct), while MARFT relies on single-turn preference optimization, our framework leverages multi-turn interactions to facilitate active error correction and ambiguity resolution. Consequently, our inference-only MAS (84.4\%) significantly outperforms the trained MARFT (78.7\%), confirming that an extended reasoning horizon contributes more to robustness than single-step alignment; AT-GRPO training further amplifies this to 87.1\%. A similar structural advantage is evident against CURE in code generation (Tab.~\ref{tab:marl_comparison_all}). While CURE generates code and unit tests in a single turn, without utilizing them for self-correction. Our framework establishes a self-refinement cycle. This enables iterative debugging using generated tests, boosting CodeContests accuracy from 22.8\% (vanilla) to 30.3\% (untrained), surpassing the CURE baseline (25.9\%), and ultimately reaching 34.2\% with training.

\revisedend

\subsection{Ablation Study}

\label{sec:ablation-study}

\begin{wrapfigure}{r}{0.5\textwidth} 
\vspace{-15pt} 
  \centering
  \includegraphics[width=0.5\textwidth]{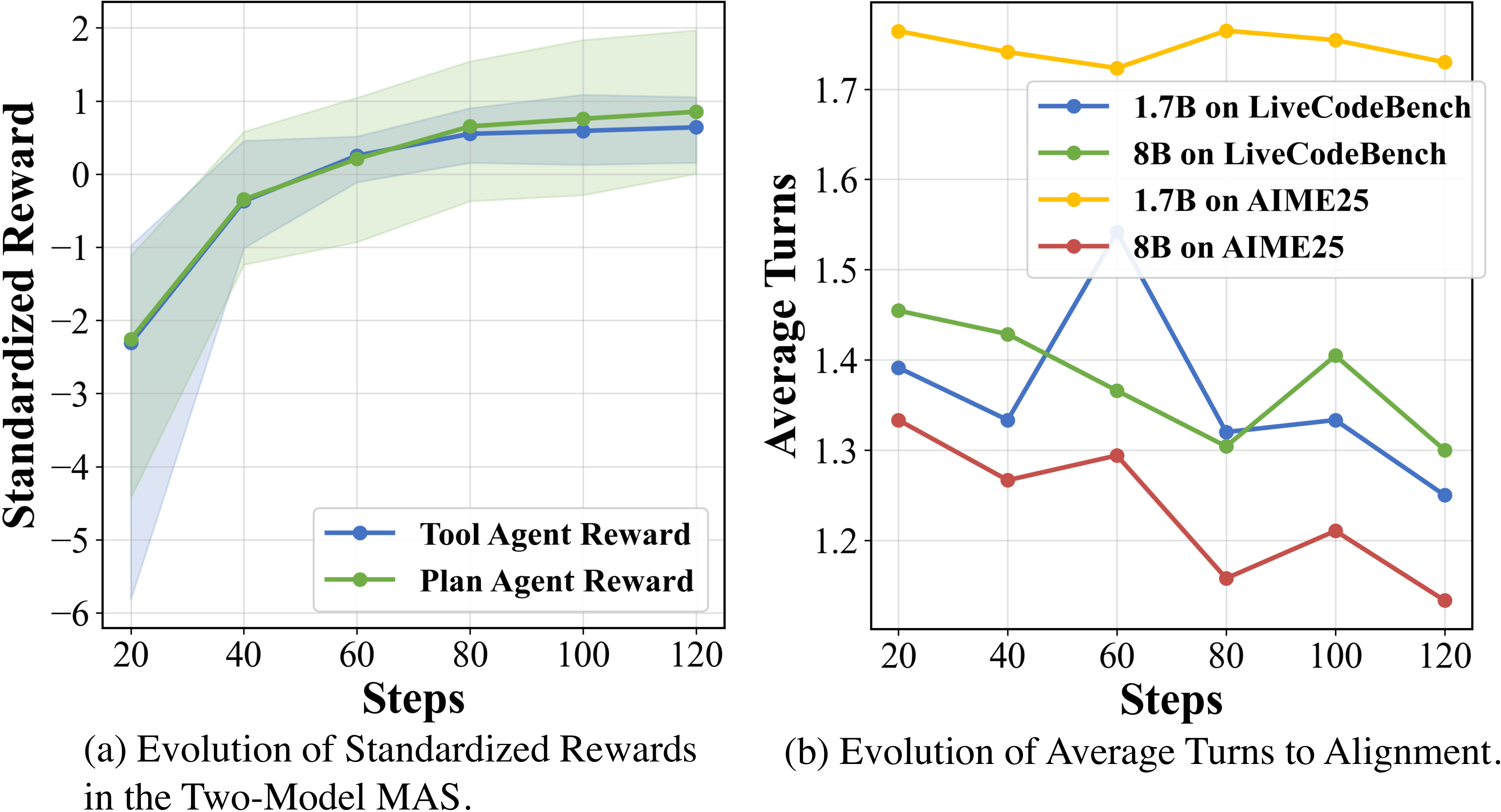} 
  \vspace{-15pt} 
  \caption{
  (a) Evolution of standardized rewards for the Tool and Plan agents in the role-specific MAS on Plan-Path with Qwen3~1.7B. Shaded bands show variability across runs. (b) Evolution of the average turns required to solve tasks. 
 }
\vspace{-15pt} 
  \label{fig:experiment} 
\end{wrapfigure}

To 
further investigate the contributions of our core training components, 
 We also conducted an ablation study with results summarized in Tab.~\ref{tab:pp-ablate} and Fig.~\ref{fig:experiment}. Our analysis yields several observations.
\begin{wraptable}{r}{0.6\linewidth} 
  \vspace{-18pt}
  \centering
  \small
  \caption{Plan-Path (Qwen3-1.7B) ablation. Performance gain $\Delta$ over the single agent baseline.}
  \label{tab:pp-ablate}
  \setlength{\tabcolsep}{1pt}    
  \renewcommand{\arraystretch}{1}
  \begin{tabular}{lcc}
    \toprule
    \textbf{Method} & \textbf{Acc.(\%)} & \(\boldsymbol{\Delta}\) \\
    \midrule
    Single agent                    & 5.00  & --     \\
    Training tool agent in SA, eval in SA  & 11.00 & +6.00 \\  
    Training code agent in SA, eval in SA  & 14.50 & +9.50 \\  
    Training in SA, eval in MAS            & 16.00 & +11.00 \\
    \midrule
    \textbf{MAS RL (role specific policies), eval in MAS} & \textbf{96.00} & \textbf{+91.00} \\
    \quad \textit{w/ Swapped Policies}                    & 6.00  & +1.00  \\
    \bottomrule
  \end{tabular}
  \vspace{-8pt}
\end{wraptable}
First, \textbf{on-policy RL training within a MAS environment is critical for effective collaboration}. As shown in Tab.~\ref{tab:pp-ablate}, training agents in a single-agent (SA) setting offers limited benefits: while individual agents improve their specialized skills (achieving 11.00 and 14.50 accuracy, respectively), their performance when combined in a MAS is only marginally better, reaching just 16.00. In stark contrast, training the agents jointly within the MAS environment boosts accuracy to 96.00. This vast performance gap demonstrates that multi-agent training is essential. It not only allows agents to co-evolve highly specialized abilities but also fosters the crucial inter-agent alignment and collaboration required for success.

Second, \textbf{RL training on MAS reinforces role-specific specialization.} We observe this across multiple metrics.  As shown in Fig.~\ref{fig:experiment}~(a) for Qwen3~1.7B on Plan-Path, the learning rewards of both the planning and tool-using agents increase throughout training, suggesting coordinated co-evolution as each adapts to the other’s improving policy. Consistent with the ablation, after training two role-specialized policies with our full method, swapping them induces a catastrophic drop from 96.0\% to 6.0\%, confirming that the agents have learned distinct and complementary functions that are not interchangeable. In our coding (LiveCodeBench) and math (AIME25) workflows, MAS interaction terminates when the two agents align (e.g., tests pass or the reasoner and tool outputs agree). Accordingly, Fig.~\ref{fig:experiment}~(b) shows that the average number of turns needed to solve a task decreases over training, providing direct evidence that the agents achieve tighter alignment and collaborate more efficiently.

\section{Conclusion}
In this paper, we proposed \textbf{AT-GRPO}, an agent- and turn-wise grouped reinforcement learning algorithm tailored for on-policy training in MAS. To support this, we introduced a novel training system capable of managing diverse MAS workflows and performing on-policy updates for multiple policies. Our extensive experiments demonstrate that our method delivers consistent gains across diverse domains. On planning tasks, it overcomes a key bottleneck of single-agent RL by boosting accuracy from a 14--47\% baseline to 96.0--99.5\%. Furthermore, it improves reasoning performance with average gains of 3.87--7.62\% on coding and 9.0--17.93\% on math tasks. Our analysis reveals that RL training in MAS context reinforces role-specific specialization, with the choice between a shared or specialized policy contingent on the task's characteristics.


\newpage
\section{Ethics Statement}

We study multi-agent reinforcement learning for large language models on planning, coding, and math tasks. Our experiments are purely computational and use public benchmarks (e.g., programmatically generated Plan-Path/Sudoku instances and widely available coding/math datasets) together with self-constructed simulators and verifiers. No human subjects, sensitive personal data, or proprietary content are involved. Code execution is performed in a sandboxed environment with restricted file I/O and no network access; tool calls are limited to deterministic checkers to prevent unintended side effects. While our methods are intended to improve reliability and sample-efficiency of agentic LLMs, we recognize dual-use risks common to autonomous systems (e.g., unsafe tool use or over-delegation). To mitigate these risks, we avoid external system operations, log all actions for auditability, and refrain from releasing any configurations that grant networked or privileged execution. We also note that base LLMs may encode societal biases that our training does not remove; results should therefore not be used for high-stakes decisions. We will release prompts, generators, and evaluation scripts to support reproducibility, subject to dataset licenses and safe-use guidelines.

\section{Reproducibility Statement}

To ensure the reproducibility of our results, we have made our datasets, code, and experimental details available. All datasets used in this study are publicly available; we provide detailed descriptions of these datasets and all data preprocessing steps in Sec.~\ref{sec:exp} and Appendix~\ref{app:training_detail}. The source code used for our experiments is included in the supplementary material. Upon acceptance, we will release the complete, documented source code under a permissive open-source license to facilitate the reproduction of all presented results. Key hyperparameters, model architectures, and training configurations are also detailed in Appendix~\ref{app:training_detail}.

\section{Use of LLM}
During the preparation of this manuscript, a large language model was utilized to aid in polishing the grammar and improving the clarity of the text. The authors reviewed and edited all outputs to ensure the final content accurately reflects our original ideas and are fully responsible for all statements and conclusions presented.

\bibliography{iclr2026_conference} 

@article{yu2025orcaloca,
  title={Orcaloca: An llm agent framework for software issue localization},
  author={Yu, Zhongming and Zhang, Hejia and Zhao, Yujie and Huang, Hanxian and Yao, Matrix and Ding, Ke and Zhao, Jishen},
  journal={arXiv preprint arXiv:2502.00350},
  year={2025}
}

@article{shao2024deepseekmath,
title={Deepseekmath: Pushing the limits of mathematical reasoning in open language models},
author={Shao, Zhihong and Wang, Peiyi and Zhu, Qihao and Xu, Runxin and Song, Junxiao and Bi, Xiao and Zhang, Haowei and Zhang, Mingchuan and Li, YK and Wu, Yang and others},
journal={arXiv preprint arXiv:2402.03300},
year={2024}
}

@article{wang2025ragen,
title={Ragen: Understanding self-evolution in llm agents via multi-turn reinforcement learning},
author={Wang, Zihan and Wang, Kangrui and Wang, Qineng and Zhang, Pingyue and Li, Linjie and Yang, Zhengyuan and Jin, Xing and Yu, Kefan and Nguyen, Minh Nhat and Liu, Licheng and others},
journal={arXiv preprint arXiv:2504.20073},
year={2025}
}

@article{feng2025group,
title={Group-in-group policy optimization for llm agent training},
author={Feng, Lang and Xue, Zhenghai and Liu, Tingcong and An, Bo},
journal={arXiv preprint arXiv:2505.10978},
year={2025}
}

@article{white2024livebench,
title   = {LiveBench: A Challenging, Contamination-Free LLM Benchmark},
author  = {White, Colin and Dooley, Samuel and Roberts, Manley and Pal, Arka and Feuer, Ben and Jain, Siddhartha and Shwartz-Ziv, Ravid and Jain, Neel and Saifullah, Khalid and Naidu, Siddartha and Hegde, Chinmay and LeCun, Yann and Goldstein, Tom and Neiswanger, Willie and Goldblum, Micah},
journal = {arXiv preprint arXiv:2406.19314},
year    = {2024},
url     = {https://arxiv.org/abs/2406.19314}
}

@misc{deepmind2024codecontests,
title        = {CodeContests},
author       = {{DeepMind}},
howpublished = {https://github.com/google-deepmind/code\_contests},
year         = {2024},
note         = {GitHub repository; archived Dec 6, 2024}
}

@article{Qian2025ToolRL,
title   = {ToolRL: Reward is All Tool Learning Needs},
author  = {Qian, Cheng and Acikgoz, Emre Can and He, Qi and Wang, Hongru and Chen, Xiusi and Hakkani-T{"u}r, Dilek and Tur, Gokhan and Ji, Heng},
journal = {arXiv preprint arXiv:2504.13958},
year    = {2025},
url     = {https://arxiv.org/abs/2504.13958}
}

@article{Liao2025MARFT,
title   = {MARFT: Multi-Agent Reinforcement Fine-Tuning},
author  = {Liao, Junwei and Wen, Muning and Wang, Jun and Zhang, Weinan},
journal = {arXiv preprint arXiv:2504.16129},
year    = {2025},
url     = {https://arxiv.org/abs/2504.16129v3}
}

@article{Hendrycks2021APPS,
  title   = {Measuring Coding Challenge Competence with APPS},
  author  = {Dan Hendrycks and Steven Basart and Saurav Kadavath and Mantas Mazeika and Andy Zou and Dawn Song and Jacob Steinhardt},
  journal = {arXiv:2105.09938},
  year    = {2021},
  url     = {https://arxiv.org/abs/2105.09938}
}

@misc{AIME2024,
  title        = {AIME 2024 Problems (AoPS Wiki)},
  author       = {{Mathematical Association of America \& AoPS Community}},
  year         = {2024},
  howpublished = {\url{https://artofproblemsolving.com/wiki/index.php/2024_AIME_I} \& \url{https://artofproblemsolving.com/wiki/index.php/2024_AIME_II_Problems}},
  note         = {Accessed 2025-09-11}
}

@misc{AIME2025,
  title        = {AIME 2025 Problems (AoPS Wiki)},
  author       = {{Mathematical Association of America \& AoPS Community}},
  year         = {2025},
  howpublished = {\url{https://artofproblemsolving.com/wiki/index.php/2025_AIME_I_Problems} \& \url{https://artofproblemsolving.com/wiki/index.php/2025_AIME_II_Problems}},
  note         = {Accessed 2025-09-11}
}

@inproceedings{He2024OlympiadBench,
  title     = {OlympiadBench: A Challenging Benchmark for Promoting AGI with Olympiad-Level Bilingual Multimodal Scientific Problems},
  author    = {Chaoqun He and Renjie Luo and Yuzhuo Bai and Shengding Hu and others},
  booktitle = {ACL},
  year      = {2024},
  url       = {https://arxiv.org/abs/2402.14008}
}

@article{Qwen3TR,
  title   = {Qwen3 Technical Report},
  author  = {An Yang and the Qwen Team},
  journal = {arXiv:2505.09388},
  year    = {2025},
  url     = {https://arxiv.org/abs/2505.09388}
}

@misc{belcak2025slm_agentic,
  title        = {Small Language Models are the Future of Agentic AI},
  author       = {Peter Belcak and Greg Heinrich and Shizhe Diao and Yonggan Fu and Xin Dong and Saurav Muralidharan and Yingyan Celine Lin and Pavlo Molchanov},
  year         = {2025},
  eprint       = {2506.02153},
  archivePrefix= {arXiv},
  primaryClass = {cs.AI},
  doi          = {10.48550/arXiv.2506.02153}
}

@inproceedings{tao2024magis,
  title     = {MAGIS: LLM-Based Multi-Agent Framework for GitHub Issue Resolution},
  author    = {Wei Tao and Yucheng Zhou and Yanlin Wang and Wenqiang Zhang and Hongyu Zhang and Yu Cheng},
  booktitle = {NeurIPS 2024},
  year      = {2024},
  eprint    = {2403.17927},
  archivePrefix = {arXiv},
  url       = {https://proceedings.neurips.cc/paper_files/paper/2024/file/5d1f02132ef51602adf07000ca5b6138-Paper-Conference.pdf}
}

@misc{liu2024drugagent,
  title        = {DrugAgent: Automating AI-aided Drug Discovery Programming through LLM Multi-Agent Collaboration},
  author       = {Shengchao Liu and others},
  year         = {2024},
  eprint       = {2411.15692},
  archivePrefix= {arXiv},
  primaryClass = {cs.AI}
}

@misc{inoue2024drugagent,
  title        = {DrugAgent: Multi-Agent Large Language Model-Based Reasoning for Drug-Target Interaction Prediction},
  author       = {Yuki Inoue and others},
  year         = {2024},
  eprint       = {2408.13378},
  archivePrefix= {arXiv},
  primaryClass = {q-bio.QM}
}

@misc{ghafarollahi2024sciagents,
  title        = {SciAgents: Automating Scientific Discovery through Multi-Agent Intelligent Graph Reasoning},
  author       = {Alireza Ghafarollahi and Markus J. Buehler},
  year         = {2024},
  eprint       = {2409.05556},
  archivePrefix= {arXiv},
  primaryClass = {cs.AI}
}

@misc{wu2023autogen,
  title        = {AutoGen: Enabling Next-Gen LLM Applications via Multi-Agent Conversation},
  author       = {Qingyun Wu and Gagan Bansal and Jieyu Zhang and Yiran Wu and Beibin Li and Erkang Zhu and Li Jiang and Xiaoyun Zhang and Shaokun Zhang and Jiale Liu and Ahmed Awadallah and Ryen W. White and Doug Burger and Chi Wang},
  year         = {2023},
  eprint       = {2308.08155},
  archivePrefix= {arXiv},
  primaryClass = {cs.CL}
}

@inproceedings{park2025maporl,
  title     = {MAPoRL: Multi-Agent Post-Co-Training for Collaborative Large Language Models with Reinforcement Learning},
  author    = {Park, Chanwoo and Han, Seungju and Guo, Xingzhi and Ozdaglar, Asuman E. and Zhang, Kaiqing and Kim, Joo-Kyung},
  booktitle = {Proceedings of the 63rd Annual Meeting of the Association for Computational Linguistics (ACL)},
  year      = {2025},
  address   = {Vienna, Austria},
  publisher = {Association for Computational Linguistics},
  url       = {https://aclanthology.org/2025.acl-long.1459/},
  doi       = {10.18653/v1/2025.acl-long.1459}
}

@misc{park2025maporl2,
  title        = {MAPoRL2: Multi-Agent Post-Co-Training for Collaborative LLMs with Reinforcement Learning},
  author       = {Park, Chanwoo and Han, Seungju and Guo, Xingzhi and Ozdaglar, Asuman E. and Zhang, Kaiqing and Kim, Joo-Kyung},
  year         = {2025},
  howpublished = {OpenReview preprint},
  url          = {https://openreview.net/pdf?id=f85TQ7hyzh},
  note         = {Influence-aware verification reward; multi-agent PPO with a learned verifier}
}

@misc{wang2025cure,
  title         = {Co-Evolving LLM Coder and Unit Tester via Reinforcement Learning},
  author        = {Wang, Yinjie and Yang, Ling and Tian, Ye and Shen, Ke and Wang, Mengdi},
  year          = {2025},
  eprint        = {2506.03136},
  archivePrefix = {arXiv},
  primaryClass  = {cs.LG},
  url           = {https://arxiv.org/abs/2506.03136}
}

@misc{chen2025mhgpo,
  title         = {Heterogeneous Group-Based Reinforcement Learning for LLM-based Multi-Agent Systems},
  author        = {Chen, Guanzhong and Yang, Shaoxiong and Li, Chao and Liu, Wei and Luan, Jian and Xu, Zenglin},
  year          = {2025},
  eprint        = {2506.02718},
  archivePrefix = {arXiv},
  primaryClass  = {cs.LG},
  url           = {https://arxiv.org/abs/2506.02718}
}

@misc{liu2025spiral,
  title         = {SPIRAL: Self-Play on Zero-Sum Games Incentivizes Reasoning via Multi-Agent Multi-Turn Reinforcement Learning},
  author        = {Liu, Bo and Guertler, Leon and Yu, Simon and Liu, Zichen and Qi, Penghui and Balcells, Daniel and Liu, Mickel and Tan, Cheston and Shi, Weiyan and Lin, Min and Lee, Wee Sun and Jaques, Natasha},
  year          = {2025},
  eprint        = {2506.24119},
  archivePrefix = {arXiv},
  primaryClass  = {cs.AI},
  doi           = {10.48550/arXiv.2506.24119},
  url           = {https://arxiv.org/abs/2506.24119}
}

@inproceedings{ma2024cory,
  title     = {Coevolving with the Other You: Fine-Tuning LLM with Sequential Cooperative Multi-Agent Reinforcement Learning},
  author    = {Ma, Hao and Hu, Tianyi and Pu, Zhiqiang and Liu, Boyin and Ai, Xiaolin and Liang, Yanyan and Chen, Min},
  booktitle = {Advances in Neural Information Processing Systems (NeurIPS)},
  year      = {2024},
  url       = {https://proceedings.neurips.cc/paper_files/paper/2024/file/1c2b1c8f7d317719a9ce32dd7386ba35-Paper-Conference.pdf}
}

@article{yao2023react,
  title={ReAct: Synergizing Reasoning and Acting in Language Models},
  author={Yao, Shunyu and Zhao, Jeffrey and Yu, Dian and Du, Nan and Shafran, Izhak and Narasimhan, Karthik and Cao, Yuan},
  journal={arXiv:2210.03629},
  year={2023},
  url={https://arxiv.org/abs/2210.03629}
}

@article{schick2023toolformer,
  title={Toolformer: Language Models Can Teach Themselves to Use Tools},
  author={Schick, Timo and Dwivedi-Yu, Jane and Dessi, Roberto and Raileanu, Roberta and Lomeli, Maria and Hambro, Eric and Zettlemoyer, Luke and Cancedda, Nicola and Scialom, Thomas},
  journal={arXiv:2302.04761},
  year={2023},
  url={https://arxiv.org/abs/2302.04761}
}

@article{shinn2023reflexion,
  title={Reflexion: Language Agents with Verbal Reinforcement Learning},
  author={Shinn, Noah and Cassano, Federico and Berman, Edward and others},
  journal={arXiv:2303.11366},
  year={2023},
  url={https://arxiv.org/abs/2303.11366}
}

@article{ahn2022saycan,
  title={Do As I Can, Not As I Say: Grounding Language in Robotic Affordances},
  author={Ahn, Michael and Brohan, Anthony and Brown, Noah and others},
  journal={arXiv:2204.01691},
  year={2022},
  url={https://arxiv.org/abs/2204.01691}
}

@article{wang2023voyager,
  title={Voyager: An Open-Ended Embodied Agent with Large Language Models},
  author={Wang, Guanzhi and Xie, Yuqi and Jiang, Yunfan and others},
  journal={arXiv:2305.16291},
  year={2023},
  url={https://arxiv.org/abs/2305.16291}
}

@article{sheng2025hybridflow,
  title={HybridFlow: A Flexible and Efficient RLHF Framework},
  author={Sheng, Guangming and Zhang, Chi and Ye, Zilingfeng and Wu, Xibin and Zhang, Wang and Zhang, Ru and Peng, Yanghua and Lin, Haibin and Wu, Chuan},
  journal={arXiv:2409.19256},
  year={2025},
  url={https://arxiv.org/abs/2409.19256}
}

@misc{volcengine2025verl,
  title={{VERL}: Volcano Engine Reinforcement Learning for LLMs},
  author={{Volcano Engine}},
  year={2025},
  howpublished={\url{https://github.com/volcengine/verl}}
}

@article{fu2025areal,
  title={{AReaL}: A Large-Scale Asynchronous Reinforcement Learning System for LLMs},
  author={Fu, W. and others},
  journal={arXiv:2505.24298},
  year={2025},
  url={https://arxiv.org/abs/2505.24298}
}

@article{Chen2025CodeSteer,
  title   = {CodeSteer: Symbolic-Augmented Language Models via Code/Text Guidance},
  author  = {Chen, Yongchao and Hao, Yilun and Liu, Yueying and Zhang, Yang and Fan, Chuchu},
  journal = {arXiv preprint arXiv:2502.04350},
  year    = {2025},
  doi     = {10.48550/arXiv.2502.04350}
}

@misc{Polaris2025,
      title={POLARIS: A Post-Training Recipe for Scaling Reinforcement Learning on Advanced Reasoning Models}, 
      url={https://hkunlp.github.io/blog/2025/Polaris},
      author={An, Chenxin and Xie, Zhihui and Li, Xiaonan and Li, Lei and Zhang, Jun and Gong, Shansan and Zhong, Ming and Xu, Jingjing and Qiu, Xipeng and Huang, Xuanjing},
      year={2025},
}

@misc{vonwerra2020trl,
  title        = {{TRL: Transformer Reinforcement Learning}},
  author       = {Leandro von Werra and Younes Belkada and Lewis Tunstall and Edward Beeching and Tristan Thrush and Nathan Lambert and Shengyi Huang and Kashif Rasul and Quentin Gallou{\'e}dec},
  year         = {2020},
  howpublished = {\url{https://github.com/huggingface/trl}},
  note         = {GitHub repository}
}

@article{sheng2024hybridflow,
  title   = {HybridFlow: A Flexible and Efficient RLHF Framework},
  author  = {Guangming Sheng and Chi Zhang and Zilingfeng Ye and Xibin Wu and Wang Zhang and Ru Zhang and Yanghua Peng and Haibin Lin and Chuan Wu},
  journal = {arXiv preprint arXiv:2409.19256},
  year    = {2024}
}

@article{hu2024openrlhf,
  title   = {OpenRLHF: An Easy-to-use, Scalable and High-performance RLHF Framework},
  author  = {Jian Hu and Xibin Wu and Zilin Zhu and Xianyu and Weixun Wang and Dehao Zhang and Yu Cao},
  journal = {arXiv preprint arXiv:2405.11143},
  year    = {2024}
}

@article{Wang2023LLMAgentsSurvey,
  title   = {A Survey on Large Language Model based Autonomous Agents},
  author  = {Wang, Liang and Liu, Qian and Song, Kun and others},
  journal = {arXiv preprint arXiv:2308.11432},
  year    = {2023},
  url     = {https://arxiv.org/abs/2308.11432}
}

@article{xi2023rise,
  title={The Rise and Potential of Large Language Model Based Agents: A Survey},
  author={Xi, Zhizheng and Chen, Wenxiang and Guo, Xin and He, Wei and Ding, Yiwen and Li, Boyang and Cui, Gongshen and Dou, Yong and Zhou, Junzhe and An, Bo and others},
  journal={arXiv preprint arXiv:2309.07864},
  year={2023}
}

@article{hu2025lmgame,
  title={lmgame-Bench: How Good are LLMs at Playing Games?},
  author={Hu, Lanxiang and Huo, Mingjia and Zhang, Yuxuan and Yu, Haoyang and Xing, Eric P and Stoica, Ion and Rosing, Tajana and Jin, Haojian and Zhang, Hao},
  journal={arXiv preprint arXiv:2505.15146},
  year={2025}
}

@inproceedings{hong2024metagpt,
  title={MetaGPT: Meta programming for a multi-agent collaborative framework},
  author={Hong, Sirui and Zhuge, Mingchen and Chen, Jonathan and Zheng, Xiawu and Cheng, Yuheng and Zhang, Ceyao and Wang, Jinlin and Wang, Zili and Yau, Steven Ka Shing and Lin, Zijuan and others},
  year={2024},
  organization={International Conference on Learning Representations, ICLR}
}

@misc{zhang2025marti,
  title        = {MARTI: A Framework for Multi-Agent LLM Systems Reinforced Training and Inference},
  author       = {Zhang, Kaiyan and Liu, Runze and Zhu, Xuekai and Tian, Kai and Zeng, Sihang and Jia, Guoli and Fan, Yuchen and Lv, Xingtai and Zuo, Yuxin and Jiang, Che and Liu, Ziyang and Wang, Jianyu and Wang, Yuru and Zhao, Ruotong and Hua, Ermo and Wang, Yibo and Wang, Shijie and Gao, Junqi and Long, Xinwei and Sun, Youbang and Ma, Zhiyuan and Cui, Ganqu and Bai, Lei and Ding, Ning and Qi, Biqing and Zhou, Bowen},
  howpublished = {\url{https://github.com/TsinghuaC3I/MARTI}},
  year         = {2025},
  note         = {Preprint}
}

@article{chen2024ioa,
  title   = {Internet of Agents: Weaving a Web of Heterogeneous Agents for Collaborative Intelligence},
  author  = {Weize Chen and Ziming You and Ran Li and Yitong Guan and Chen Qian and Chenyang Zhao and Cheng Yang and Ruobing Xie and Zhiyuan Liu and Maosong Sun},
  journal = {arXiv preprint arXiv:2407.07061},
  year    = {2024}
}

@article{wang2024moa,
  title   = {Mixture-of-Agents Enhances Large Language Model Capabilities},
  author  = {Junlin Wang and Jue Wang and Ben Athiwaratkun and Ce Zhang and James Zou},
  journal = {arXiv preprint arXiv:2406.04692},
  year    = {2024}
}

@article{ye2025xmas,
  title   = {X-MAS: Towards Building Multi-Agent Systems with Heterogeneous LLMs},
  author  = {Rui Ye and Xiangrui Liu and Qimin Wu and Xianghe Pang and Zhenfei Yin and Lei Bai and Siheng Chen},
  journal = {arXiv preprint arXiv:2505.16997},
  year    = {2025}
}

@article{pan2025whydomasfail,
  title   = {Why Do Multi-Agent LLM Systems Fail?},
  author  = {Mert Cemri and Melissa Z. Pan and Shuyi Yang and Lakshya A. Agrawal and Bhavya Chopra and Rishabh Tiwari and Kurt Keutzer and Aditya Parameswaran and Dan Klein and Kannan Ramchandran and Matei Zaharia and Joseph E. Gonzalez and Ion Stoica},
  journal = {arXiv preprint arXiv:2503.13657},
  year    = {2025}
}

@article{guo2024llmma,
  title   = {Large Language Model based Multi-Agents: A Survey of Progress and Challenges},
  author  = {Taicheng Guo and Xiuying Chen and Yaqi Wang and Ruidi Chang and Shichao Pei and Nitesh V. Chawla and Olaf Wiest and Xiangliang Zhang},
  journal = {arXiv preprint arXiv:2402.01680},
  year    = {2024}
}

@article{Mao2020RewardDesignMARL,
  title   = {Reward Design in Cooperative Multi-agent Reinforcement Learning for Packet Routing},
  author  = {Hangyu Mao and Zhibo Gong and Zhen Xiao},
  journal = {arXiv preprint arXiv:2003.03433},
  year    = {2020},
  url     = {https://arxiv.org/abs/2003.03433}
}

@inproceedings{Sheikh2020DEMADDPG,
  title     = {Multi-Agent Reinforcement Learning for Problems with Combined Individual and Team Reward},
  author    = {Hassam Ullah Sheikh and Ladislau B{\"o}l{\"o}ni},
  booktitle = {Proceedings of the International Joint Conference on Neural Networks (IJCNN)},
  year      = {2020},
  url       = {https://arxiv.org/abs/2003.10598}
}

@misc{chen2025setsleveragingselfverificationselfcorrection,
      title={SETS: Leveraging Self-Verification and Self-Correction for Improved Test-Time Scaling}, 
      author={Jiefeng Chen and Jie Ren and Xinyun Chen and Chengrun Yang and Ruoxi Sun and Jinsung Yoon and Sercan Ö Arık},
      year={2025},
      eprint={2501.19306},
      archivePrefix={arXiv},
      primaryClass={cs.AI},
      url={https://arxiv.org/abs/2501.19306}, 
}

@misc{cobbe2021training,
      title={Training Verifiers to Solve Math Word Problems}, 
      author={Karl Cobbe and Vineet Kosaraju and Mohammad Bavarian and Mark Chen and Heewoo Jun and Lukasz Kaiser and Matthias Plappert and Jerry Tworek and Jacob Hilton and Reiichiro Nakano and Christopher Hesse and John Schulman},
      year={2021},
      eprint={2110.14168},
      archivePrefix={arXiv},
      primaryClass={cs.LG}
}

@book{sutton2018rl,
  title        = {Reinforcement Learning: An Introduction},
  author       = {Sutton, Richard S. and Barto, Andrew G.},
  edition      = {2nd},
  year         = {2018},
  publisher    = {MIT Press}
}
\bibliographystyle{plainnat} 

\appendix
\newpage
\appendix

\section{Extended Comparison of RL Training for LLM-based MAS}
\label{app:mas-rl-comparison}
\revisedbegin
To complement the discussion on RL training for LLM-based multi-agent systems in
Sec.~\ref{sec:relatedwork}, we provide an extended, axis-by-axis comparison of
representative frameworks in Table~\ref{tab:mas_rl_comparison}.  We contrast
MAPoRL, MARFT, MARTI, CURE, and our method along several key design
dimensions: (i) whether agents share a single policy or use role-specific
policies; (ii) whether the interaction pattern is sequential, parallel, or a
hybrid of both; (iii) whether the framework supports multi-turn interaction;
(iv) whether agent roles are heterogeneous; (v) the number of evaluation task
domains; and (vi) whether the underlying RL algorithm is designed as a
general-purpose MAS training scheme rather than being tightly coupled to a
single task.

\begin{table}[htbp]
  
  \small
  \centering
  \caption{Comparison of RL-based LLM multi-agent training frameworks.}
  \label{tab:mas_rl_comparison}
  \resizebox{\linewidth}{!}{%
  \revisedbegin
  \begin{tabular}{lcccccc}
    \toprule
    \textbf{Method} 
      & \textbf{Policy sharing} 
      & \textbf{Execution pattern} 
      & \textbf{Multi-turn} 
      & \textbf{Role heterogeneity} 
      & \textbf{$\geq 2$  domains} verification 
      & \textbf{Generic MAS RL algo.} \\
    \midrule
    MAPoRL~\cite{park2025maporl} 
      & R 
      & P 
      & \cmark 
      & \xmark 
      & \xmark 
      & \xmark \\
    MARFT~\cite{Liao2025MARFT} 
      & R 
      & S 
      & \xmark 
      & \cmark 
      & \xmark 
      & \cmark \\
    MARTI~\cite{zhang2025marti} 
      & R 
      & S+P 
      & \cmark 
      & \cmark 
      & \xmark 
      & \xmark \\
    CURE~\cite{wang2025cure} 
      & S 
      & P 
      & \xmark 
      & \cmark 
      & \xmark 
      & \xmark \\
    \textbf{Ours (StrongerMAS)} 
      & S+R 
      & S+P 
      & \cmark 
      & \cmark 
      & \cmark 
      & \cmark \\
    \bottomrule
  \end{tabular}%
  
\revisedend

  }
  \revisedbegin
   \begin{minipage}{0.95\linewidth}
    \footnotesize\emph{Note.} For MARFT~\citep{Liao2025MARFT}, we report the
    characteristics of version v3 of the framework, corresponding to the
    preprint available prior to the completion of this work.
  \end{minipage}
  \revisedend
\end{table}

\revisedend
As summarized in Tab.~\ref{tab:mas_rl_comparison}, \revised{StrongerMAS} is the only RL-based LLM multi-agent training framework that simultaneously supports both shared and role-specific policies, hybrid sequential–parallel execution, multi-turn interaction, and heterogeneous agent roles, while being validated on \emph{multiple} task domains and implemented as a general-purpose MAS RL algorithm. In contrast, MAPoRL~\citep{park2025maporl} and CURE~\citep{wang2025cure} restrict policies to a purely role-shared setting and are tailored to specific tasks, MARFT~\cite{Liao2025MARFT}
restricts agents to single-turn sequential interactions, and MARTI~\cite{zhang2025marti} merely introduces basic single-agent RL algorithms (e.g., GRPO) to the MAS setting. . This combination of flexible workflow expressivity (S+R, S+P, heterogeneous roles) and broad, cross-domain evaluation makes \revised{StrongerMAS} a more faithful and scalable abstraction for training cooperative LLM-based MAS.

\section{Reward Design}
\label{app:reward}
\subsection{Math reward design}

We consider math QA with horizon $T$ and optional tool calls.
Let $h_t$ be the dialogue/tool history at turn $t$.
We adopt \textsc{Math-Verifier}\footnote{\textsc{Math-Verify} (Hugging Face), GitHub: \href{https://github.com/huggingface/Math-Verify}{huggingface/Math-Verify}. We use it as a parsing/normalization front-end and then apply a numeric comparator.} as the checker front-end.

Define a numeric comparator with tolerance $\delta$:
\[
\textsc{NumEq}_{\delta}(a,b) \;=\; \mathbf{1}\!\left\{|a-b|\le \delta\ \ \text{or}\ \ \frac{|a-b|}{\max(1,|b|)}\le \delta\right\},
\quad \delta{=}10^{-6}\,.
\]
\deleted{
$
\lambda_{\mathrm{math}} = 0.70.$}

\paragraph{Team reward.}
Sparse pass at termination via numerical equality.
We broadcast the same scalar reward to all turns:
\[
r^{\mathrm{team}}_t \;=\; \mathbf{1}\{\textsc{CheckFinal}_{\textsc{MathVerifier}{+}\textsc{NumEq}}(h){=}\textsf{pass}\}\in\{0,1\},
\qquad \forall t.
\]

\paragraph{Local rewards.}
Each agent $i$ at turn $t$ uses a masked convex combination of verifiable sub-scores
$s^{i}_{\ell,t}\in[0,1]$:
\[
r^{\mathrm{loc}}_{t,i}
\;=\;
m_{t,i}
\sum_{\ell\in\{\mathrm{fmt},\mathrm{tool},\mathrm{step}\}}
c^{i}_{\ell}\, s^{i}_{\ell,t},
\qquad
\sum_{\ell} c^{i}_{\ell} = 1,
\]
where $m_{t,i}\in\{0,1\}$ is a verifiability mask.

\paragraph{Reasoner local design.}
Coefficients:
\[
c^{\mathrm{Reasoner}}_{\mathrm{fmt}} = 0.20,\quad
c^{\mathrm{Reasoner}}_{\mathrm{tool}} = 0.00,\quad
c^{\mathrm{Reasoner}}_{\mathrm{step}} = 0.80.
\]
Component scores (pure numerical check):
\[
s^{\mathrm{Reasoner}}_{\mathrm{fmt},t}
= \mathbf{1}\{\text{required output schema matched at turn }t\},
\]
\[
s^{\mathrm{Reasoner}}_{\mathrm{step},t} =
\begin{cases}
\textsc{NumEq}_{\delta}\!\big(\hat y,\, y^\star\big), & \text{if } \textsc{MathVerifier} \text{ extracts a numeric } \hat y,\\[2pt]
0, & \text{otherwise},
\end{cases}
\]
and we do not use a tool-related step, so the corresponding score is implicitly
$s^{\mathrm{Reasoner}}_{\mathrm{tool},t}\equiv 0$.
The mask is
\[
m^{\mathrm{Reasoner}}_t
= \mathbf{1}\{y^\star\ \text{available (}\textsc{MathVerifier}\text{) at turn }t\}.
\]
For the Reasoner, $r^{\mathrm{loc}}_{t,\mathrm{Reasoner}}$ is obtained by plugging these
coefficients, scores and the mask into the generic form above.

\subsection{Code reward design}

We consider code synthesis with unit tests.
Let $\mathcal{S}$ be the active test suite and
\[
p=\frac{1}{|\mathcal{S}|}\sum_{t\in\mathcal{S}}\mathbf{1}\{\textsc{Run}(t,\text{code})=\textsf{pass}\}\in[0,1].
\]
Team reward is dense; we again broadcast it over turns:
\[
r^{\mathrm{team}}_t \;=\; p,
\qquad \forall t.
\]

\paragraph{Local rewards.}
Each agent $i$ at turn $t$ uses a masked convex combination of verifiable sub-scores
$s^{i}_{\ell,t}\in[0,1]$:
\[
r^{\mathrm{loc}}_{t,i}
\;=\;
m_{t,i}
\sum_{\ell}
c^{i}_{\ell}\, s^{i}_{\ell,t},
\qquad
\sum_{\ell} c^{i}_{\ell} = 1.
\]

\paragraph{Coder local reward.}
We define the Coder's local reward as a weighted combination of basic sanity checks and
the fraction of golden tests passed by the generated code.
Coefficients:
\[
c^{\mathrm{Coder}}_{\mathrm{build}} = 0.10,\quad
c^{\mathrm{Coder}}_{\mathrm{run}} = 0.10,\quad
c^{\mathrm{Coder}}_{\mathrm{nr}} = 0.80.
\]
Let $\mathcal{T}^{\text{gold}}$ be the fixed set of golden unit tests for this problem.
Component scores at turn $t$ are
\[
s^{\mathrm{Coder}}_{\mathrm{build},t}
  = \mathbf{1}\{\text{the candidate code compiles/imports without syntax errors at }t\},
\]
\[
s^{\mathrm{Coder}}_{\mathrm{run},t}
  = \mathbf{1}\{\text{a smoke subset of }\mathcal{T}^{\text{gold}}\text{ runs without uncaught exceptions/timeout at }t\},
\]
\[
s^{\mathrm{Coder}}_{\mathrm{nr},t}
  = \frac{1}{|\mathcal{T}^{\text{gold}}|}
    \sum_{u \in \mathcal{T}^{\text{gold}}}
    \mathbf{1}\{\textsc{Run}(u,\text{code}) = \textsf{pass}\},
\]
i.e., the fraction of golden tests passed by the current code $\text{code}$.
We apply an availability mask
\[
m^{\mathrm{Coder}}_t
= \mathbf{1}\{\text{build/run logs and golden-test results are available at }t\},
\]
and define the Coder's local reward as
\[
r^{\mathrm{loc}}_{t,\mathrm{Coder}}
= m^{\mathrm{Coder}}_t \Big(
    c^{\mathrm{Coder}}_{\mathrm{build}} s^{\mathrm{Coder}}_{\mathrm{build},t}
  + c^{\mathrm{Coder}}_{\mathrm{run}}   s^{\mathrm{Coder}}_{\mathrm{run},t}
  + c^{\mathrm{Coder}}_{\mathrm{nr}}    s^{\mathrm{Coder}}_{\mathrm{nr},t}
  \Big).
\]

\paragraph{Tester local design.}

Coefficients :
\[
c^{\mathrm{Tester}}_{\mathrm{valid}} = 0.20,\quad c^{\mathrm{Tester}}_{\mathrm{nr}} = 0.80.
\]
\cancel{$c^{\mathrm{Tester}}_{\mathrm{cov}} = 0.80.$}

Component scores:
\[
s^{\mathrm{Tester}}_{\mathrm{valid},t}
= \mathbf{1}\{\text{new/edited tests are executable, deterministic, and respect I/O at }t\},
\]

\[
\cancel{%
s^{\mathrm{Tester}}_{\mathrm{cov}} =
\begin{cases}
\min\!\left(1,\ \dfrac{\mathrm{MutScore}}{\tau_{\mathrm{mut}}}\right), & \text{mutation analysis available},\\[6pt]
0, & \text{otherwise},
\end{cases}
}
\]

\deleted{
where $(x)_{+}=\max(x,0)$, $\mathrm{MutScore}_k\in[0,1]$ is the mutation score on golden code, $\mathrm{BrCov}_k\in[0,1]$ is branch coverage, and thresholds are fixed as
$
\tau_{\mathrm{mut}}=0.60,\tau_{\mathrm{cov}}=0.10.
$}

\revisedbegin
\[
s^{\mathrm{Tester}}_{\mathrm{nr},t}
  = \frac{1}{|\mathcal{U}|}
    \sum_{u \in \mathcal{U}}
    \mathbf{1}\{\textsc{Run}(u,\text{code}^\star) = \textsf{pass}>\tau_{mut}\},
\]
where $
\tau_{\mathrm{mut}}=0.60$.
\revisedend

Mask:
\[
m^{\mathrm{Tester}}_t
= \mathbf{1}\{\text{test runner and mutation/coverage reports available at }t\}.
\]
The Tester local reward $r^{\mathrm{loc}}_{t,\mathrm{Tester}}$ is obtained by combining
$c^{\mathrm{Tester}}_{\cdot}$, $s^{\mathrm{Tester}}_{\cdot,t}$ and $m^{\mathrm{Tester}}_t$
via the generic local-reward formula.

\subsection{Sudoku reward design}

We consider $N{\times}N$ Sudoku. Let $h_t$ be the answer action at turn $t$ and
$\textsc{Solved}(\cdot)$ check row/column/subgrid validity.
Team reward is a sparse success signal at termination, broadcast across turns:
\[
r^{\mathrm{team}}_t\;=\; \mathbf{1}\{\ \textsc{Solved}(h){=}\textsf{true}\}\in\{0,1\},
\qquad \forall t.
\]
\deleted{
We set the team--local mixing coefficient to a fixed number
\[
\lambda_{\mathrm{sudoku}} = 0.60.
\]
For each agent $i\in\{\mathrm{Reasoner},\mathrm{Tool}\}$ at turn $k$, with verifiability mask $m^i_k\in\{0,1\}$, the per-agent learning reward is $r^i_k \;=\; \lambda_{\mathrm{sudoku}}\, r^{\mathrm{team}}_k \;+\; (1-\lambda_{\mathrm{sudoku}})\, m^i_k\, r^{i,\mathrm{loc}}_k.$}

\paragraph{Local rewards.}
Each agent $i$ at turn $t$ uses a masked convex combination of verifiable sub-scores
$s^{i}_{\ell,t}\in[0,1]$:
\[
r^{\mathrm{loc}}_{t,i}
\;=\;
m_{t,i}
\sum_{\ell}
c^{i}_{\ell}\, s^{i}_{\ell,t},
\qquad
\sum_{\ell} c^{i}_{\ell} = 1.
\]

\paragraph{Reasoner local design.}
Coefficients :
\[
c^{\mathrm{Reasoner}}_{\mathrm{fmt}} = 0.1,\quad
c^{\mathrm{Reasoner}}_{\mathrm{legal}} = 0.1,\quad
c^{\mathrm{Reasoner}}_{\mathrm{prog}} = 0.80.
\]
Component scores (let $G_t$ be the current grid, $G_{t-1}$ the previous grid; $0$ denotes empty):
\[
s^{\mathrm{Reasoner}}_{\mathrm{fmt},t} = \mathbf{1}\{\text{action format is valid (full }N{\times}N\text{ grid or list of }[r,c,v])\},
\]
\[
s^{\mathrm{Reasoner}}_{\mathrm{legal},t} = \mathbf{1}\{\text{no row/column/subgrid duplicates in }G_t\},
\]
\[
s^{\mathrm{Reasoner}}_{\mathrm{prog},t} =
\frac{1}{N^2}\sum_{r,c}\mathbf{1}\{G_{t-1}[r,c]{=}0,\ G_t[r,c]{\neq}0\}.
\]
Mask:
\[
m^{\mathrm{Reasoner}}_t
= \mathbf{1}\{\text{we can parse the action and compute legality/progress at }t\}.
\]

\paragraph{Tool (executor) local design.}
Coefficients (fixed):
\[
c^{\mathrm{Tool}}_{\mathrm{fmt}} = 0.10,\quad
c^{\mathrm{Tool}}_{\mathrm{exec}} = 0.10,\quad
c^{\mathrm{Tool}}_{\mathrm{san}} = 0.80.
\]
Component scores:
\[
s^{\mathrm{Tool}}_{\mathrm{fmt},t} = \mathbf{1}\{\text{API/schema valid; values in }[1,N];\ \text{indices in bounds}\},
\]
\[
s^{\mathrm{Tool}}_{\mathrm{exec},t} = \mathbf{1}\{\text{no runtime error/timeout when applying edits}\},
\]
\[
s^{\mathrm{Tool}}_{\mathrm{san},t} =
\begin{cases}
1, & \text{if all applied edits satisfy local Sudoku constraints},\\
0, & \text{otherwise}.
\end{cases}
\]
Mask:
\[
m^{\mathrm{Tool}}_t
= \mathbf{1}\{\text{executor logs available and legality checks computed at }t\}.
\]

\subsection{Plan-Path reward design}

We consider 2D grid path planning on a $H{\times}W$ map with horizon $T$ and four-neighborhood moves.
Let $d_t$ be the Manhattan distance from the current position to the goal at turn $t$ and
$d_0=\max(1,\text{initial distance})$ for normalization.
Team reward is dense and distance-improving:
\[
r^{\mathrm{team}}_t \;=\;
\begin{cases}
1, & \text{if at goal at }t,\\[4pt]
\max\!\bigl(0,\ (d_{t-1}-d_t)/d_0\bigr), & \text{otherwise}.
\end{cases}
\]
\deleted{
We set the team--local mixing coefficient to a fixed number$\lambda_{\mathrm{plan}} = 0.50.$
For each agent $i\in\{\mathrm{Planner},\mathrm{Tool}\}$ with mask $m^i_k\in\{0,1\}$,
$
r^i_k \;=\; \lambda_{\mathrm{plan}}\, r^{\mathrm{team}}_k \;+\; (1-\lambda_{\mathrm{plan}})\, m^i_k\, r^{i,\mathrm{loc}}_k.
$
}

Local rewards are masked convex combinations
\[
r^{\mathrm{loc}}_{t,i}
\;=\;
m_{t,i}
\sum_{\ell} c^{i}_{\ell}\, s^{i}_{\ell,t},
\qquad \sum_{\ell} c^{i}_{\ell}=1.
\]

\paragraph{Planner local design.}
Coefficients (fixed):
\[
c^{\mathrm{Planner}}_{\mathrm{fmt}} = 0.10,\quad
c^{\mathrm{Planner}}_{\mathrm{leg}} = 0.10,\quad
c^{\mathrm{Planner}}_{\mathrm{sp}} = 0.80.
\]
Component scores at turn $t$ (action $a_t\in\{\mathrm{U,D,L,R}\}$; $\mathcal{N}$ denotes passable neighbors; $\textsc{SPNext}$ is 1 if $a_t$ lies on at least one shortest path from $s_{t-1}$ to goal, else 0):
\[
s^{\mathrm{Planner}}_{\mathrm{fmt},t} = \mathbf{1}\{a_t\in\{\mathrm{U,D,L,R}\}\},
\]
\[
s^{\mathrm{Planner}}_{\mathrm{leg},t} = \mathbf{1}\{\text{next cell in-bounds and not a wall}\},
\]
\[
s^{\mathrm{Planner}}_{\mathrm{sp},t} =
\begin{cases}
1, & \text{if }\textsc{SPNext}(a_t){=}1,\\
0, & \text{otherwise}.
\end{cases}
\]
Mask:
\[
m^{\mathrm{Planner}}_t = \mathbf{1}\{\text{map known and shortest-path oracle available at }t\}.
\]

\paragraph{Tool (executor/simulator) local design.}
Coefficients:
\[
c^{\mathrm{Tool}}_{\mathrm{fmt}} = 0.10,\quad
c^{\mathrm{Tool}}_{\mathrm{exec}} = 0.10,\quad
c^{\mathrm{Tool}}_{\mathrm{shape}} = 0.80.
\]
Component scores (let $\phi_t=-d_t$ be the potential used in shaping):
\[
s^{\mathrm{Tool}}_{\mathrm{fmt},t} = \mathbf{1}\{\text{action list parsable as }[\texttt{``U",``D",``L",``R"}]\},
\]
\[
s^{\mathrm{Tool}}_{\mathrm{exec},t} = \mathbf{1}\{\text{no invalid move applied; simulation advances}\},
\]
\[
s^{\mathrm{Tool}}_{\mathrm{shape},t} =
\mathbf{1}\{\phi_t \ge \phi_{t-1}\},
\]
i.e., the potential does not decrease.
Mask:
\[
m^{\mathrm{Tool}}_t = \mathbf{1}\{\text{execution logs and potentials }(\phi_{t-1},\phi_t)\text{ available}\}.
\]

\subsection{Sokoban reward design}

We consider Sokoban with horizon $T$ on a fixed grid.
Let $B$ be the number of boxes and $b_t$ the number of boxes on goal at turn $t$.
Team reward is dense in box-on-goal ratio with terminal success at completion:
\[
r^{\mathrm{team}}_t \;=\;
\begin{cases}
1, & \text{if all boxes on goals at }t,\\[4pt]
b_t/B, & \text{otherwise}.
\end{cases}
\]
\deleted{
We set the team--local mixing coefficient to a fixed number
\[
\lambda_{\mathrm{sok}} = 0.40.
\]
For each agent $i\in\{\mathrm{Planner},\mathrm{Tool}\}$ with mask $m^i_t\in\{0,1\}$,
\[
r^i_t \;=\; \lambda_{\mathrm{sok}}\, r^{\mathrm{team}}_t \;+\; (1-\lambda_{\mathrm{sok}})\, m^i_t\, r^{\mathrm{loc}}_{t,i}.
\]}

Local rewards are masked convex combinations
\[
r^{\mathrm{loc}}_{t,i}
\;=\;
m_{t,i}
\sum_{\ell} c^{i}_{\ell}\, s^{i}_{\ell,t},
\qquad \sum_{\ell} c^{i}_{\ell}=1.
\]

\paragraph{Planner local design.}
Coefficients (fixed):
\[
c^{\mathrm{Planner}}_{\mathrm{fmt}} = 0.10,\quad
c^{\mathrm{Planner}}_{\mathrm{leg}} = 0.10,\quad
c^{\mathrm{Planner}}_{\mathrm{dlk}} = 0.80.
\]
Component scores at turn $t$ (action $a_t\in\{\mathrm{U,D,L,R}\}$; $\textsc{PushOK}=1$ if a planned push does not collide and stays in-bounds; $\textsc{DeadlockFree}=1$ if the move avoids standard static corner deadlocks for boxes not on goals):
\[
s^{\mathrm{Planner}}_{\mathrm{fmt},t} = \mathbf{1}\{a_t\in\{\mathrm{U,D,L,R}\}\},
\]
\[
s^{\mathrm{Planner}}_{\mathrm{leg},t} = \mathbf{1}\{\text{step is in-bounds and not into wall; if pushing, }\textsc{PushOK}=1\},
\]
\[
s^{\mathrm{Planner}}_{\mathrm{dlk},t} =
\begin{cases}
1, & \text{if }\textsc{DeadlockFree}=1,\\
0, & \text{otherwise}.
\end{cases}
\]
Mask:
\[
m^{\mathrm{Planner}}_t = \mathbf{1}\{\text{grid known and deadlock heuristics evaluable at }t\}.
\]

\paragraph{Tool (executor/simulator) local design.}
Coefficients (fixed):
\[
c^{\mathrm{Tool}}_{\mathrm{fmt}} = 0.10,\quad
c^{\mathrm{Tool}}_{\mathrm{exec}} = 0.10,\quad
c^{\mathrm{Tool}}_{\mathrm{pot}} = 0.80.
\]
Let $\psi_t = - \sum_{x\in \text{boxes}} \min_{g\in \text{goals}} \bigl(|x_r-g_r|+|x_c-g_c|\bigr)$ be the box-to-goal potential (larger is better).
Component scores:
\[
s^{\mathrm{Tool}}_{\mathrm{fmt},t} = \mathbf{1}\{\text{action list parsable; symbols match } \{\mathrm{U,D,L,R}\}\},
\]
\[
s^{\mathrm{Tool}}_{\mathrm{exec},t} = \mathbf{1}\{\text{no illegal push; no wall/box collision}\},
\]
\[
s^{\mathrm{Tool}}_{\mathrm{pot},t} = \mathbf{1}\{\psi_t \ge \psi_{t-1}\}.
\]
Mask:
\[
m^{\mathrm{Tool}}_t = \mathbf{1}\{\text{execution logs and potentials }(\psi_{t-1},\psi_t)\text{ available}\}.
\]

\revised{\subsection{Outcome-only reward design}}
\label{app:reward-outcome}
\revisedbegin
The shaped rewards in Sections~\ref{app:reward} 1--5 provide rich, task-specific feedback
(e.g., shortest-path signals in Plan-Path and deadlock heuristics in Sokoban).
To isolate the contribution of such complex shaping from that of the
AT-GRPO algorithm itself, we additionally consider a simplified
\textbf{outcome-only} reward design used in our ablation studies.

The \textbf{team reward} is strictly binary and episodic. Let 
$\mathbb{I}(\text{Success})$ 
denote the environment success indicator. The team reward is defined and
broadcast over turns as
\[
r^{\mathrm{team}}_t \;=\; \mathbb{I}(\text{Success}),
\qquad \forall t.
\]

The \textbf{per-agent local reward} in this setting is an auxiliary
binary signal that only checks whether agent $i$ produced a validly
formatted action (e.g., correct API call or JSON structure). Let
$\mathbb{I}(\text{FmtValid}^i_t)$ denote the indicator that the output
of agent $i$ at turn $t$ satisfies all formatting constraints. We define
\[
r^{\mathrm{loc}}_{t,i}
\;=\;
r^{i,\mathrm{out}}_t
\;=\;
\mathbb{I}\big(\text{FmtValid}^i_t\big).
\]

The final per-agent reward $r_{t,i}$ is then combined according to Eq.~\ref{eq:reward},
\[
r_{t,i} \;=\; \alpha\, r^{\mathrm{team}}_t \;+\; r^{\mathrm{loc}}_{t,i},
\]
where we use a fixed, task-independent $\alpha$ shared with the shaped-reward
configurations. By utilizing sparse episodic rewards and simple formatting checks
rather than dense shaping signals, this outcome-only configuration is significantly
more general and provides a baseline for our algorithm ablation.
\revisedend

\subsection{Theoretical Justification for Greedy Turn-Level Transitions}

In this section, we formally justify the optimality of greedy selection based on environment-verified rewards. Consider the underlying MDP with optimal action-value function $Q^*(s,a)$. The Bellman optimality principle implies that any policy $\pi^*$ satisfying $\pi^*(s) \in \arg\max_a Q^*(s,a)$ is optimal \citep{sutton2018rl}.

We operate in a setting where the environment returns an \emph{outcome-based verifiable reward}, $r_{\mathrm{ver}}(s,a)$, for each action. We posit that this reward acts as a monotonic proxy for the true value function $Q^*(s,a)$: a higher verification score directly corresponds to a higher probability of final success. Consequently, maximizing the immediate verifiable reward is structurally equivalent to maximizing the long-term optimal value. We formalize this alignment as follows:

\begin{assumption}[Monotonicity of Verification Feedback]
\label{ass:outcome-aligned}
For any state $s$ and actions $a_1, a_2$, the verifiable reward preserves the ordering of the optimal action-value function:
\begin{equation}
    r_{\mathrm{ver}}(s,a_1) > r_{\mathrm{ver}}(s,a_2) \implies Q^*(s,a_1) \ge Q^*(s,a_2).
\end{equation}
This implies that $r_{\mathrm{ver}}(s, \cdot)$ and $Q^*(s, \cdot)$ induce consistent rankings over the action space at any state $s$.
\end{assumption}

\begin{lemma}[Equivalence of Maximizers]
\label{lem:greedy-equivalence}
Under Assumption~\ref{ass:outcome-aligned}, the set of actions maximizing the verifiable reward is a subset of the actions maximizing the optimal $Q$-function:
\begin{equation}
    \arg\max_{a} r_{\mathrm{ver}}(s,a) \subseteq \arg\max_{a} Q^*(s,a).
\end{equation}
\end{lemma}

\begin{proof}
Let $a^*_{\mathrm{ver}} \in \arg\max_a r_{\mathrm{ver}}(s,a)$. Suppose, for the sake of contradiction, that there exists an action $a'$ such that $Q^*(s,a') > Q^*(s,a^*_{\mathrm{ver}})$. By the contrapositive of Assumption~\ref{ass:outcome-aligned}, strict inequality in $Q^*$ implies strict inequality in $r_{\mathrm{ver}}$ (given consistent rankings), which would imply $r_{\mathrm{ver}}(s,a') \ge r_{\mathrm{ver}}(s,a^*_{\mathrm{ver}})$.
Since $a^*_{\mathrm{ver}}$ is a maximizer, strict inequality is impossible. If equality holds, $a'$ is also a maximizer of $r_{\mathrm{ver}}$, and by the consistency assumption, it must share the same optimal $Q$-value. Thus, any action maximizing $r_{\mathrm{ver}}$ necessarily maximizes $Q^*$.
\end{proof}

\begin{proposition}[Optimality of Verifier-Greedy Policy]
\label{prop:greedy-optimal}
Let $\pi_{\mathrm{ver}}$ be a deterministic policy such that $\pi_{\mathrm{ver}}(s) \in \arg\max_a r_{\mathrm{ver}}(s,a)$ for all states $s$. Under Assumption~\ref{ass:outcome-aligned}, $\pi_{\mathrm{ver}}$ is an optimal policy.
\end{proposition}

\begin{proof}
By Lemma~\ref{lem:greedy-equivalence}, selecting an action that maximizes the immediate verification score ensures that $\pi_{\mathrm{ver}}(s) \in \arg\max_a Q^*(s,a)$. Consequently, $\pi_{\mathrm{ver}}$ satisfies the Bellman optimality equation at every state.
\end{proof}

In our implementation, we approximate this policy by sampling candidate actions and greedily selecting the one with the highest $r_{\mathrm{ver}}$. Proposition~\ref{prop:greedy-optimal} guarantees that this strategy effectively performs a greedy search over the support of sampled actions with respect to the true optimal value function $Q^*$, avoiding the myopic bias typically associated with greedy transitions.

\section{Experiment Details}
\subsection{Training Details}
\label{app:training_detail}

All methods share the same hyperparameters unless noted. The maximum response length is \textbf{4096} tokens, and the (task-specific) maximum prompt length is set to accommodate turn-by-turn dialogue history: \textbf{8192} tokens for \emph{mathematics} and \emph{code} tasks, and \textbf{16384} tokens for all other symbolic tasks. Training uses a global batch size of \textbf{128}, with \textbf{PPO mini-batch size 64} and gradient clipping at \textbf{1.0}. The actor is optimized with Adam at a learning rate of \textbf{1e-6} and weight decay \textbf{0.01}. We adopt \textbf{GRPO} for advantage estimation with $\gamma{=}1.0$ and $\lambda{=}1.0$. Entropy regularization is off ($\texttt{entropy\_coeff}{=}0$). The sample temperature $T_{sample}=1.0$, top-$p{=}1.0$, top-$k{=}-1$, and 4 sample per prompt; validation is deterministic (temperature \textbf{0}, \texttt{do\_sample}=\textbf{False}). rewards are computed by a rule-based function (\texttt{compute\_score}) when provided. Both models are trained for 150 steps.

\subsection{Prompt Design}
\label{app:prompt}
\definecolor{codercolor}{HTML}{EBF5FB}  
\definecolor{testercolor}{HTML}{E8F6F3} 
\definecolor{coderborder}{HTML}{5499C7}
\definecolor{testerborder}{HTML}{48C9B0}

\newtcolorbox{promptbox}[3]{
  breakable,
  colback=#2,
  colframe=#3,
  fonttitle=\bfseries,
  title=#1,
  arc=2mm,
  boxrule=1pt,
  left=4mm, right=4mm, top=3mm, bottom=3mm,
}

\newcommand{\placeholder}[1]{\texttt{\{#1\}}}
\newcommand{\outputformat}[1]{\texttt{<#1>}}

\textbf{Code MAS Workflow}

\subsection*{Phase 1: Generation}
In the initial phase, both agents are given a problem description. The Coder is prompted to generate a solution, while the Tester is prompted to generate a corresponding test case.

\begin{promptbox}{Code Agent (Coder): Turn 0}{codercolor}{coderborder}
\textbf{Input:}
\begin{itemize}
    \item \textbf{Problem:} A natural language description of a programming task.
\end{itemize}

\textbf{Prompt:}

    You are a helpful assistant that writes Python to solve the problem. 
    Think step by step, then output code.
    Important:
    - Read all inputs via input().
    - Print all results with print().
    - Do not hardcode or fabricate inputs.    
    Now solve:
    Problem: ```problem description```
    First, decide on the number and types of inputs required (e.g.,
    x = int(input()), b = int(input())), then implement the solution
    and print the result.
    Please answer in the following format: 
    Code: \verb|`|\verb|`|\verb|`|python (your code here)\verb|`|\verb|`|\verb|`|

\textbf{Output:} Code
\end{promptbox}

\begin{promptbox}{Unit Tester Agent (Test-Case Author): Turn 0}{testercolor}{testerborder}
\textbf{Input:}
\begin{itemize}
    \item \textbf{Problem:} A natural language description of a programming task, e.g., \placeholder{problem}.
\end{itemize}

\textbf{Prompt:}

    You are a helpful assistant that creates unit test cases (input +
expected output) for a coding task.

Problem: ``` problem discrption```

Provide one new high-quality test case. Before giving the test case,
reason carefully to ensure the output is correct, then derive the
output for your chosen input.
Respond in the format:
 **Test Input:**\verb|`|\verb|`|\verb|`|input here\verb|`|\verb|`|\verb|`| **Test Output:**\verb|`|\verb|`|\verb|`|output here\verb|`|\verb|`|\verb|`|

\textbf{Output:} Test input, Test Output.

\end{promptbox}

\subsection*{Phase 2: Refinement}
In subsequent turns, the agents receive feedback based on mismatches between the generated code and test cases. They are prompted to refine their previous outputs.

\begin{promptbox}{Code Agent (Coder): Turn $>$ 0}{codercolor}{coderborder}
\textbf{Input:}
\begin{itemize}
    \item \textbf{Problem:} The original problem description, \placeholder{problem}.
    \item \textbf{Mismatch History:} A record of previous code, test inputs, expected outputs, and actual execution outputs, highlighting any differences, \placeholder{mismatch\_history}.

\end{itemize}

\textbf{Prompt:}

You are a helpful assistant that corrects and refines code.

Important:
- Read inputs via input(); output with print().
- Do not hardcode inputs.

Problem: \{problem\}

Use the history below to guide your fixes:

\{mismatch\_history\}

If your previous code crashed, first fix the bug.

If execution succeeded but outputs mismatched the expected output,
decide if the test case is correct.
- If the test is correct, refine your code to pass it.
- If the test is wrong, verify your program's logic and keep it.

Provide the final, corrected code. Respond in the format:

Code: \verb|`|\verb|`|\verb|`|python \# your code here\verb|`|\verb|`|\verb|`|

\textbf{Output:} Code
\end{promptbox}

\begin{promptbox}{Unit Tester Agent (Test-Case Author): Turn $>$ 0}{testercolor}{testerborder}
\textbf{Input:}
\begin{itemize}
    \item \textbf{Problem:} The original problem description, \placeholder{problem}.
    \item \textbf{Mismatch History:} A record showing the test case and the differing execution output from the Coder's program, \placeholder{mismatch\_history}.

\end{itemize}

\textbf{Prompt:}
You are an assistant that checks and refines unit tests for a
coding task.

Problem: {problem}

Analyze the history below:

\{mismatch\_history\}

First, decide whether your previous test case was correct (watch for
misunderstandings of the task). If it was wrong or unclear, provide a
corrected test case.
Respond in the format:

     **Test Input:**,
     **Test Output:**

\textbf{Output:} Test input, test output.

\end{promptbox}

\textbf{Math MAS Workflow}

\definecolor{reasonercolor}{HTML}{FFFDE7}  
\definecolor{reasonerborder}{HTML}{F9A825} 

\subsection*{Phase 1: Generation}
In the initial phase, two complementary agents are given the same math problem. The \emph{Reasoning Agent} produces a step-by-step mathematical solution and a boxed final answer. The \emph{Python Tool Agent} writes executable Python that computes (and prints) the final answer.

\begin{promptbox}{Reasoning Agent: Turn 0}{reasonercolor}{reasonerborder}
\textbf{Input:}
\begin{itemize}
\item \textbf{Problem:} A mathematical problem in natural language.
\end{itemize}

\textbf{Prompt:}

You are a helpful assistant that solves math problems via careful reasoning.

Problem:
{problem}

First, outline the key reasoning steps. Then carry out the full solution.
After solving, present the final answer in a LaTeX box.

Before giving the full reasoning, summarize the steps clearly in:
**Reasoning Steps:**
`reasoning steps here`

Then provide your complete solution concisely. Put your final answer in:
\#\#\#

Rules:

* The boxed value must be a single number or expression (simplified if possible).
* Do not add words after the box; only the final value goes after \#\#\#\#.

Output format:

1. Your reasoning (short and clear).
2. Final line must contain only the boxed answer, e.g., \#\#\#\# 123.

\textbf{Output:} Reasoning solution and a final answer after \#\#\#\#.
\end{promptbox}

\begin{promptbox}{Python Tool Agent (Coder for Math): Turn 0}{codercolor}{coderborder}
\textbf{Input:}
\begin{itemize}
\item \textbf{Problem:} The same mathematical problem, \placeholder{problem}.
\end{itemize}

\textbf{Prompt:}

You are a helpful programming assistant that writes Python to solve the math problem.

\paragraph{Problem}
\texttt{\{problem\}}

\paragraph{Requirements}
\begin{itemize}
  \item Write correct, readable Python that computes the final answer.
  \item Think step by step in comments if helpful.
  \item Use only the standard library and deterministic math (no internet, no randomness).
  \item At the end, PRINT ONLY the final numeric or symbolic answer (nothing else).
\end{itemize}

\textbf{Output:} Code (the program prints the final answer).
\end{promptbox}

\subsection*{Phase 2: Refinement}
From the second turn onward, agents receive feedback derived from mismatches between the Reasoning Agent’s boxed answer and the Python Tool Agent’s printed output. Each agent uses the history to refine its output.

\begin{promptbox}{Reasoning Agent (Math Solver): Turn $>$ 0}{reasonercolor}{reasonerborder}
\textbf{Input:}
\begin{itemize}
\item \textbf{Problem:} The original problem, \placeholder{problem}.
\item \textbf{Mismatch History:} Prior reasoning (\placeholder{reasoning\_solution}), its extracted answer (\placeholder{reasoning\_extracted\_answer}), the Python code (\placeholder{code\_solution}), and the code’s printed output (\placeholder{code\_extracted\_answer}), summarized as \placeholder{mismatch\_history}.
\end{itemize}

\textbf{Prompt:}

You are a helpful assistant that refines mathematical solutions through reasoning.

Problem: problem

History (previous attempts and outputs):
mismatch history

First, compare your previous boxed answer with the Python Tool Agent’s printed output.

* If the code output corrects a computational slip in your reasoning, adopt the corrected value.
* If the code likely has a bug (e.g., mishandled edge cases, precision, domains), 
keep the mathematically correct answer and explain briefly.

Then solve the problem again, more robustly.

Before giving the full reasoning, summarize the key steps clearly:
**Reasoning Steps:**
`reasoning steps here`

Finish with the final answer after:
\#\#\#\#

Final line must contain only the boxed value (no extra text).

\textbf{Output:} Updated reasoning and a final answer after \#\#\#\#.
\end{promptbox}

\begin{promptbox}{Python Tool Agent (Coder for Math): Turn $>$ 0}{codercolor}{coderborder}
\textbf{Input:}
\begin{itemize}
\item \textbf{Problem:} The original problem, \placeholder{problem}.
\item \textbf{Mismatch History:} Prior code and printed output, and the Reasoning Agent’s solution and boxed answer, summarized as \placeholder{mismatch\_history}.
\end{itemize}

\textbf{Prompt:}

You are a helpful programming assistant that refines Python solutions for math problems.

Problem:
problem

History (reasoning vs. execution mismatches):
mismatch history

Tasks:

1. Judge whether the Reasoning Agent’s boxed answer or your previous printed result
   is more likely correct (consider numerical stability, edge cases, exact vs. float).
2. Fix or rewrite the code so it reliably computes the correct final answer.

   * Prefer exact arithmetic (fractions, integers, rational simplification) when possible.
   * Add minimal checks for domain/edge cases.
   * Keep outputs deterministic.

Respond in the format:

**Code:**

\verb|`|\verb|`|\verb|`|python
\# corrected code here
\# print ONLY the final answer on the last line
\verb|`|\verb|`|\verb|`|

\textbf{Output:} Refined code (the program prints the final answer).
\end{promptbox}

\textbf{Sudoku MAS Workflow}

In the initial phase, two complementary agents are given the same Sudoku-solving task on an \(n{\times}n\) grid. The \emph{Tool Agent} writes executable Python that outputs either a completed grid or a list of fill steps. The \emph{Plan Agent} inspects the task, the tool code, and its execution output, then decides the final solution.

\definecolor{toolcolor}{HTML}{F1F8FF}
\definecolor{toolborder}{HTML}{1E3A8A}
\definecolor{plancolor}{HTML}{FFF8E1}
\definecolor{planborder}{HTML}{B45309}

\lstset{
  basicstyle=\ttfamily\small,
  frame=single,
  framerule=0.4pt,
  columns=fullflexible,
  keepspaces=true,
  showstringspaces=false,
  breaklines=true
}

\begin{promptbox}{Tool Agent (Sudoku Coder)}{toolcolor}{toolborder}
\textbf{Input:}
\begin{itemize}
\item \textbf{Task Description:} \placeholder{task}, including grid size, rules (rows/columns/sub-grids contain unique digits), and any constraints.
\item \textbf{Env Context:} \placeholder{env\_context} (e.g., \placeholder{size}, \placeholder{subgrid\_size}, \placeholder{puzzle}, \placeholder{observation}).
\end{itemize}

\textbf{Prompt:}

You are an AI assistant designed to be helpful. Utilize your programming
expertise to address the task. Propose Python code (within a single python
code block) for the user to run. Ensure each response contains only ONE code
block. Use the 'print' function to output EITHER:
  (A) the completed grid as a JSON array of arrays, OR
  (B) a JSON list of fill steps (r,c,v) using 1-based indices.

Formatting requirements:

* The program's output is the Sudoku solution:
    eg: [[5,3,4,6,7,8,9,1,2], ..., [3,4,5,2,8,6,1,7,9]]
  
* Print ONLY the JSON (no extra text, no comments).

Task:
Solve the sizexsize Sudoku. Fill digits 1..size ; rows, columns, and
sub-grids must have unique digits.

Current puzzle (dots denote blanks):
observation

Environment:
- size
- subgrid\_size: subgrid\_size
- notes/constraints: constraints

\textbf{Output:} Code (program prints either the completed grid JSON or a JSON list of fill steps).
\end{promptbox}

\begin{promptbox}{Plan Agent (Planner \& Verifier)}{plancolor}{planborder}
\textbf{Input:}
\begin{itemize}
\item \textbf{Task Description:} \placeholder{task}.
\item \textbf{Tool Code:} \placeholder{tool\_code}.
\item \textbf{Tool Execution Output:} \placeholder{tool\_execution\_output}.
\item \textbf{Tool Proposed Solution:} \placeholder{tool\_solution} (JSON grid or JSON steps).
\item \textbf{Observation (for reference):} \placeholder{observation}.
\end{itemize}

\textbf{Prompt:}
You are a planning and reasoning agent. You will receive:

* The original task description
* The Tool Agent's code
* The code execution output (a JSON grid or JSON steps)

Your job is to reason carefully, decide the final Sudoku solution, and format
your response EXACTLY as specified.

Instructions:

* Read the task, inspect the code, and verify the execution output against the
  Sudoku rules: rows, columns, and sub-grids must contain unique digits in 1..n.
* If the tool's output is a complete, valid solution, adopt it.
* If it is incomplete or violates constraints, correct it or provide your own.
* Keep reasoning concise but explicit: explain why the final result is valid.

FORMATTING IS MANDATORY.
Give the final answer AFTER the line that begins with \#\#\#\#.
You may return EITHER:
  - a completed grid as JSON, OR
  - a JSON list of fill steps (r,c,v), 1-based indices.

Examples:

\#\#\#\# [[5,3,4,6,7,8,9,1,2], ..., [3,4,5,2,8,6,1,7,9]]

\#\#\#\# [[1,3,4],[2,1,6],[9,9,1]]

\textbf{Output:} Final Sudoku answer (completed grid JSON or JSON steps).
\end{promptbox}
\section{Plan-Path MAS Workflow}

\subsection*{Phase 1: Generation}
In the initial phase, two complementary agents are given the same path-planning task on a grid/world. The \emph{Tool Agent} writes executable Python that outputs an action list (e.g., $[U,R,D,L]$). The \emph{Plan Agent} inspects the task, the tool code, and its execution output, then decides the final action list.
\definecolor{toolcolor}{HTML}{F1F8FF}
\definecolor{toolborder}{HTML}{1E3A8A} 
\begin{promptbox}{Tool Agent (Path Coder): Turn 0}{toolcolor}{toolborder}
\textbf{Input:}
\begin{itemize}
\item \textbf{Task Description:} \placeholder{task}, including grid/map, start, goal, obstacles, and constraints.
\item \textbf{Env Context:} \placeholder{env\_context} (e.g., \placeholder{grid}, \placeholder{start}, \placeholder{goal}, \placeholder{obstacles}, \placeholder{constraints}).
\end{itemize}

\textbf{Prompt:}

You are an AI assistant designed to be helpful. Utilize your programming
expertise to address the task. Propose Python code (within a single python
code block) for the user to run. Ensure each response contains only ONE code
block. Use the 'print' function to output the action list that moves from the
start to the goal. You may output the full action list if you can reach the
target, or a partial list if uncertain.

Formatting requirements:

* Begin the Python block with \`python and end with \`.
* The program's output IS the action list (e.g., [U,R,D,L]).
* Print ONLY the action list (no extra text).

Task:
{task}

Environment:
{env\_context}

\textbf{Output:} Code (program prints an action list, e.g., \texttt{ [U,R,D,L]}).
\end{promptbox}

\begin{promptbox}{Plan Agent (Planner \& Verifier): Turn 0}{plancolor}{planborder}
\textbf{Input:}
\begin{itemize}
\item \textbf{Task Description:} \placeholder{task}.
\item \textbf{Tool Code:} \placeholder{tool\_code}.
\item \textbf{Tool Execution Output:} \placeholder{tool\_execution\_output}.
\item \textbf{Tool Proposed Action:} \placeholder{tool\_action}.
\end{itemize}

\textbf{Prompt:}

You are a planning and reasoning agent. You will receive:

* The original task description
* The Code Agent’s (Tool Agent’s) code
* The code execution output

Your job is to reason carefully, decide the final action list, and format your
response EXACTLY as specified.

Instructions:

* Read the task, inspect the code, and verify the execution output against the
  task requirements and environment constraints (bounds, obstacles, goal).
* If the code/output is correct and sufficient, adopt it.
* Otherwise, improve or override it with your own reasoning.
* Keep reasoning concise but explicit: justify why the final action is correct.

FORMATTING IS MANDATORY.
Give the final action list AFTER the line that begins with \#\#\#\#.
Example:

\#\#\#\# [U,R,D,L]

\textbf{Output:} Final action list.
\end{promptbox}

\subsection*{Phase 2: Refinement}
From the second turn onward, agents receive feedback based on mismatches between the Tool Agent’s printed action list and feasibility checks from the environment or the Plan Agent’s assessment. Each agent uses the history to refine its output.

\begin{promptbox}{Tool Agent (Path Coder): Turn $>$ 0}{toolcolor}{toolborder}
\textbf{Input:}
\begin{itemize}
\item \textbf{Task Description:} \placeholder{task}.
\item \textbf{Mismatch/Trajectory History:} Prior code and printed actions, planner feedback, and (action, state) pairs, summarized as \placeholder{action\_state\_history}.
\end{itemize}

\textbf{Prompt:}

Refine your Python solution to produce a correct, executable action list.

Task:
{task}

History (previous attempts, planner feedback, and trajectory):
{action\_state\_history}

Requirements:

* Output must be an action list that reaches the goal without violating
  constraints (stay in-bounds, avoid obstacles).
* If certain, print the full list; if uncertain, print a safe partial prefix.
* Single python code block only; program output IS the action list.
* Begin with `python and end with `; print ONLY the action list (e.g., [U,R,D,L]).

Respond in the format:

**Code:**

\verb|`|\verb|`|\verb|`| python

\# corrected code here
\# last line prints ONLY the action list

\verb|`|\verb|`|\verb|`|

\textbf{Output:} Refined code (program prints an action list).
\end{promptbox}

\begin{promptbox}{Plan Agent (Planner \& Verifier): Turn $>$ 0}{plancolor}{planborder}
\textbf{Input:}
\begin{itemize}
\item \textbf{Task Description:} \placeholder{task}.
\item \textbf{Tool Code:} \placeholder{tool\_code}.
\item \textbf{Tool Execution Output:} \placeholder{tool\_execution\_output}.
\item \textbf{Tool Proposed Action:} \placeholder{tool\_action}.
\item \textbf{Action State History (if any):} For each step $i$, The $i$-th action is $a_i$. The $i$-th state is $s_i$. Summarized as action state history.
\end{itemize}

\textbf{Prompt:}

You are a planning and reasoning agent.

Task:
{task}

Tool Agent’s latest code and output:

* Code: {tool\_code}
* Execution output: {tool\_execution\_output}
* Proposed action: {tool\_action}

Trajectory/history:
{action\_state\_history}

Instructions:

* Verify feasibility of the proposed action sequence step by step.
* If it collides, goes out of bounds, loops, or fails to reach the goal,
  correct it (you may shorten, extend, or replace the sequence).
* Prefer the simplest valid plan; if uncertain, provide the best safe prefix
  and explain briefly.
* Keep reasoning concise but explicit.

FORMATTING IS MANDATORY.
Give the FINAL action list AFTER the line that begins with \#\#\#\#.
Example:

\#\#\#\# [U,R,D,L]

\textbf{Output:} Final action list.
\end{promptbox}

\revised{\section{Ablation study of Outcome reward}}
\revisedbegin
\begin{table}[htbp]
    \centering
    \revised{\caption{\textbf{Performance Comparison with Sparse Outcome-Only Rewards.} To address concerns regarding reward engineering, we evaluate \textsc{AT-GRPO} using only sparse outcome signals (Outcome-only), removing all intermediate heuristics. Even without dense guidance, our method maintains high performance and significantly outperforms the baselines.}}
    \label{tab:outcome_ablation}
    \revised{
    \resizebox{0.7\linewidth}{!}{
    \begin{tabular}{l|cc|cc|c}
        \toprule
        \multirow{2}{*}{\textbf{Task}} & \multicolumn{2}{c|}{\textbf{Baselines}} & \multicolumn{2}{c|}{\textbf{AT-GRPO (Ours)}} & \textbf{Robustness} \\
        & SA & MAS & \textbf{Outcome-only} & Dense (Original) & (Drop $\Delta$) \\
        \midrule
        Sokoban   & 48.0\% & 72.0\% & \textbf{93.0\%} & 96.0\% & \textcolor{gray}{-3.0\%} \\
        Sudoku    & 9.0\%  & 16.0\% & \textbf{99.5\%} & 99.5\% & \textbf{0.0\%} \\
        Plan-Path & 12.0\% & 71.0\% & \textbf{89.0\%} & 93.0\% & \textcolor{gray}{-4.0\%} \\
        \bottomrule
    \end{tabular}
    }}
\end{table}

A potential concern with the dense task-specific rewards (detailed in Appendix~\ref{app:reward}) is that they might provide ``oracle'' guidance (e.g., distance-to-goal heuristics in Plan-Path), thereby simplifying the reasoning challenge. To disentangle the contribution of the \textsc{AT-GRPO} algorithm from the reward design, we evaluate our method using the \textbf{Outcome-only} reward formulation defined in Appendix~\ref{app:reward-outcome}. In this setting, all intermediate heuristic signals are removed, and the agents receive positive feedback only upon successfully solving the final task, exactly matching the sparse signal availability of the baselines.

Table~\ref{tab:outcome_ablation} compares the performance of \textsc{AT-GRPO} under dense versus sparse outcome-only rewards against the SA and MAS baselines. We observe two key findings:
\begin{enumerate}
    \item \textbf{Independence from Dense Heuristics:} The removal of dense rewards results in only marginal performance degradation. For instance, on the \textit{Plan-Path} task—where the dense reward provided shortest-path information—the accuracy drops by only 4.0\% (from 93.0\% to 89.0\%). On \textit{Sudoku}, the performance remains identical at 99.5\%. This indicates that while dense rewards accelerate learning, they are not a prerequisite for the model's success.
    
    \item \textbf{Superiority over Baselines:} Even in the sparse outcome-only setting, \textsc{AT-GRPO} maintains a decisive advantage over the baselines. On \textit{Plan-Path}, our sparse-reward performance (89.0\%) vastly outperforms the SA baseline (12.0\%) and the MAS baseline (71.0\%). This dramatic gap ($+77\%$ vs. SA) under identical reward conditions strongly refutes the hypothesis that our results are confounded by reward engineering. Instead, it demonstrates that the cooperative group optimization mechanism is intrinsically capable of solving complex planning tasks without oracle guidance.
\end{enumerate}
\revisedend

\revisedbegin

\section{Multi turn single agent}
\label{sec:samt}
\begin{table}[htbp]
\centering
\small
\caption{Single-agent ablations on \textbf{Code} and \textbf{Math} (Qwen3 1.7B).}
\label{tab:sa-code-math-1p7b}
\begin{tabular}{lccc|ccc}
\toprule
& \multicolumn{3}{c|}{\textbf{Code}} & \multicolumn{3}{c}{\textbf{Math}} \\
\cmidrule(lr){2-4}\cmidrule(lr){5-7}
\textbf{Setting} 
& LiveCodeBench & APPS & CodeContests 
& AIME24 & AIME25 & Olympiad \\
\midrule
SA, single turn 
& 11.6 & 16.2 & 3.6 
& 13.4 & 9.8 & 22.2 \\

SA + multi-turn 
& 10.4 & 10.4 & 0.0 
& 3.3 & 6.7 & 15.8 \\

SA, single turn + RL
& 18.8 & 17.0 & 3.0 
& 10.0 & 6.7 & 23.8 \\

SA, multi-turn +RL
& 17.7 & 13.3 & 1.2 
& 6.67 & 3.3 & 16.9 \\
\bottomrule
\end{tabular}
\end{table}

\begin{table}[htbp]
\centering
\small

\caption{Single-agent ablations on \textbf{Code} and \textbf{Math} (Qwen3 8B).}
\label{tab:sa-code-math-8b}
\begin{tabular}{lccc|ccc}

\toprule
& \multicolumn{3}{c|}{\textbf{Code}} & \multicolumn{3}{c}{\textbf{Math}} \\
\cmidrule(lr){2-4}\cmidrule(lr){5-7}
\textbf{Setting} 
& LiveCodeBench & APPS & CodeContests 
& AIME24 & AIME25 & Olympiad \\
\midrule
SA, single turn 
& 22.8 & 30.2 & 15.75 
& 18.3 & 20.0 & 55.0 \\

SA + multi-turn 
& 7.8 & 20.3 & 5.12 
& 16.7 & 16.7 & 53.4 \\

SA, single turn + RL
& 25.7 & 37.0 & 12.12 
& 18.3 & 26.67 & 54.8 \\

SA, multi-turn + RL
& 16.8 & 35.4 & 11.1 
& 16.7 & 23.3 & 51.2 \\
\bottomrule
\end{tabular}
\end{table}

\paragraph{On the effectiveness of multi-turn single-agent variants.}
Tab.~\ref{tab:sa-code-math-1p7b} and Tab.~\ref{tab:sa-code-math-8b} report single-agent
ablations on Code and Math. 
For both 1.7B and 8B models, introducing a multi-turn SA variant 
(i.e., letting one agent repeatedly revise its own answer) brings no consistent
benefit over the standard single-turn SA baseline and often degrades performance, which is align with the obeservation in \cite{chen2025setsleveragingselfverificationselfcorrection} .
For example, at the SFT stage on Qwen3-1.7B, 
LiveCodeBench drops from $11.6$ to $10.4$ and AIME24 from $13.4$ to $3.3$ when
switching from single-turn SA to multi-turn SA, with similar trends on AIME25
and Olympiad. 
After RL, the single-turn SA policy still outperforms its multi-turn
counterpart across most Code and Math benchmarks for both model scales.
These results support our claim in the main text: in the absence of additional
environmental signals or feedback from complementary roles, multi-turn SA
interaction is a contrived use of extra turns that departs from the QA-style
pretraining regime and fails to translate into improved task performance, in
contrast to our multi-agent workflows where multi-turn interaction with
structured cross-agent feedback yields clear gains.

\section{System Complexity of Agent- and Turn-wise Grouping}
\label{app:complexity}

Our on-policy RL framework operates by alternating between two distinct phases: \textbf{inference} (rollout generation) and \textbf{training} (loss computation and parameter updates).
In this section, we analyze the computational and memory complexity of AT-GRPO (Alg.~\ref{alg:hatgrpo_mas}) and discuss how it scales with the number of agents and the turn horizon under different MAS interaction patterns and system constraints.

\paragraph{Notation.}
Let $N$ be the number of agents, $T$ the turn horizon, $E$ the number of
parallel environment instances, and $K$ the sampling factor (number of
candidate actions per agent--turn in tree sampling).
Let $L$ denote the average number of generated tokens per action.

\subsection{Inference time Complexity}
\subsubsection{System Design with Asynchronous vLLM Generation}

Our implementation uses a vLLM-style \emph{asynchronous} engine with continuous batching for both rollouts and evaluation: each agent--turn query $(e,i,t)$ is submitted as an independent request, and the engine maintains a token-level scheduler that dynamically adds new sequences and removes finished ones. Compared to a naive synchronous design that forms a fixed batch of agent responses and waits for the longest sequence to finish, this asynchronous scheme largely eliminates long-tail stragglers, keeps the GPUs close to saturation, and naturally interleaves agent--turns from parallel and sequential MAS workflows into efficient pipeline. 

\subsubsection{Inference-time Complexity}

During inference ($K{=}1$), we analyze the per-episode wall-clock latency. The complexity depends on the execution schedule—Sequential or Parallel—determined by the interaction logic.
Crucially, the baseline Single-Agent (SA) complexity also varies by task: for \textbf{Code} and \textbf{Math}, the SA baseline is typically single-turn ($T{=}1$), whereas for \textbf{Plan} and \textbf{Game}, the SA baseline involves multi-turn interactions ($T{>}1$). We denote the baseline latency as $\text{Time}_{\text{infer}}^{\text{SA}}$.

\paragraph{Sequential MAS.}
In this setting (e.g., \textbf{Plan}, \textbf{Game}), agents act serially within each turn to condition on updated history.
While the SA baseline requires $T$ sequential steps, the Sequential MAS requires $N$ serial agent moves for each of the $T$ turns, resulting in a critical path of $N \cdot T$.
Comparing this to the multi-turn SA baseline:
\[
\frac{\text{Time}_{\text{infer}}^{\text{Seq}}}{\text{Time}_{\text{infer}}^{\text{SA}}} 
\;\le\; 
\frac{N \cdot T}{T}
\;=\;
N.
\]
Thus, the latency overhead scales linearly with the number of agents $N$.

\paragraph{Parallel MAS.}
In this setting (e.g., \textbf{Code}, \textbf{Math}), we employ multi-round debate where all $N$ agents act in parallel in each round.
By leveraging continuous batching, the $N$ concurrent queries are processed together on the inference engine.
However, the parallelism is not unbounded: for a fixed model and hardware budget, there exists a maximum number of concurrent sequences the engine can hold in memory.

Let $B_{\max}$ denote the maximum number of concurrent sequences that can be served by the cluster (determined by GPU memory, model size, and the target context length).
With $E$ parallel environments and $K$ candidates per agent (e.g., $K$ GRPO samples), the number of sequences per MAS step is $E \cdot N \cdot K$.
To keep all agents truly parallel, we must satisfy
\[
E \cdot N \cdot K \;\le\; B_{\max}.
\]
Equivalently, the maximum parallelizable agent count is
\[
N_{\max} \;=\; \biggl\lfloor \frac{B_{\max}}{E \cdot K} \biggr\rfloor.
\]
When $N \le N_{\max}$, the $N$ agents at each turn can be fully batched, and the latency scales primarily with the debate depth $T$:
\[
\frac{\text{Time}_{\text{infer}}^{\text{Para}}}{\text{Time}_{\text{infer}}^{\text{SA}}}
\;\lesssim\;
T.
\]
When $N > N_{\max}$, the engine automatically schedules the $N$ agents in
$\lceil N / N_{\max} \rceil$ waves, and the latency bound becomes
\[
\frac{\text{Time}_{\text{infer}}^{\text{Para}}}{\text{Time}_{\text{infer}}^{\text{SA}}}
\;\lesssim\;
T \cdot \bigl\lceil N / N_{\max} \bigr\rceil,
\]
which smoothly reduces to the single-wave case when $N \le N_{\max}$.

\subsection{Training-time Complexity}

The computational bottleneck during training lies in the forward and backward passes for the collected candidate actions. For $E$ environments, $N$ agents, and $T$ turns, with $K$ samples each, the total rollout size is $|\mathcal{D}|_{\text{MAS}} = E \cdot N \cdot T \cdot K$.

The proposed agent- and turn-wise grouping introduces only \textbf{a lightweight hashing overhead} of $O(|\mathcal{D}|_{\text{MAS}})$, which is negligible compared to the token-level model execution $O(|\mathcal{D}|_{\text{MAS}} \cdot L \cdot C_{\text{model}})$.
Therefore, the complexity relationship between our multi-agent approach and the standard single-agent GRPO ($|\mathcal{D}|_{\text{SA}} = E \cdot T \cdot K$) is defined by the ratio of their rollout sizes:
\[
\frac{\text{Time}_{\text{train}}^{\text{MAS}}}{\text{Time}_{\text{train}}^{\text{SA}}} 
\;\le
\frac{|\mathcal{D}|_{\text{MAS}}}{|\mathcal{D}|_{\text{SA}}} 
\;=\; NT.
\]
This demonstrates that our method introduces no extra asymptotic complexity beyond a linear scaling with the number of agents $N$.
\subsection{Empirical Latency Study}

We conducted latency profiling on a cluster of four H100 GPUs with an effective decoding batch size of $32 \times 8$.
For the \textbf{Code} task, one on-policy iteration for the single-agent baseline ($N{=}1, T{=}1$) requires approximately 4 minutes for rollout (inference) and 1 minute for AT-GRPO training; thus, inference dominates roughly 80\% of the total wall-clock time.
Scaling to the MAS setting ($N{=}2$, multi-turn) approximately results in 8 minutes for rollout and 2 minutes for training.

In the \textbf{Game} domain, while training costs remain comparable to the Code task, we observe an inversion in inference latency. The single-agent baseline averages 2.8 minutes per rollout, whereas the MAS inference time drops to 1.5 minutes.
This reduction is attributable to the superior performance of MAS: the group efficiently completes tasks in fewer turns (triggering early termination), whereas the single-agent policy frequently struggles and exhausts the maximum turn horizon.
\section{Case Studies of MAS Workflows}
\label{app:casestudy}

This appendix presents two concrete multi-agent case studies, one in a
box-pushing grid game and one in code generation with unit tests.  For
each domain, we include the original prompts and agent-facing messages,
and we distinguish erroneous behaviors from successful ones using
\bad{} and \good{}, respectively.

\subsection{MAS for Game}
\label{app:casestudy-game}

\paragraph{Task.}
Task: Planner proposes the next action sequence; Executor calls
environment tools (simulator, legality checker, shortest-path/BFS
helper) to apply actions and return effects/observations (updated grid,
agent/box poses, success/failure flags). Episode ends when the goal is
met (all boxes on targets) or the turn budget is reached.

\paragraph{Before RL (\bad).}
Before RL: The Plan Agent gets a valid path for the box from Tool agent
but completely misses the point. It tries to follow the box's path
itself, runs straight into a wall, and fails instantly. It doesn't
understand its job is to push the box, not be the box.

\paragraph{After on-policy RL in MAS (\good).}
After on-policy RL in MAS: RL teaches the agent the difference. It
learns that rewards come from moving the box along the designated path.
This insight forces it to discover the correct low-level strategy:
first, navigate behind the box, then execute the push.

\begin{promptbox}{The Puzzle (Game Grid)}{envcolor}{envborder}
\textbf{Input (environment prompt):}

\medskip
\noindent
\texttt{\#\#\#\#\#\#\#\#\#\#\#}\\
\texttt{\#  .  .  .  .  .  .  . G  .  \#}\\
\texttt{\#  .\#\#\#  .  .  .\#  .  \#}\\
\texttt{\#  .  \#  .  .  .  .  .  \#  .  \#}\\
\texttt{\#  .  \#  .  .  .  .  .  \#  .  \#}\\
\texttt{\#  .  \#  .  B  .  .  .  \#  .  \#}\\
\texttt{\#A  .  .  .  .  .  .  \#  .  \#}\\
\texttt{\#\#\#\#\#\#\#\#\#\#\#}

\medskip
\noindent
\texttt{\#} wall,\ \texttt{.} free,\ \texttt{A} agent,\ \texttt{B} box,\ \texttt{G} goal

\medskip
\noindent
Start: A(8,1), B(6,6), G(1,8) --- 1-indexed (row,col).

\textbf{Output:} Initial grid state for the MAS.
\end{promptbox}

\begin{promptbox}{Tool Agent (Box Path via BFS)}{toolcolor}{toolborder}
\textbf{Input:}
\begin{itemize}
    \item \textbf{Task Description:} Given the grid and object positions,
    compute an optimal path for the \emph{box} from its start to the goal.
\end{itemize}

\textbf{Prompt (Tool Agent behavior):}

A tool agent perceives obstacles and calculates an optimal path for the
\textbf{box} using BFS:

\textbf{Output (Tool Agent response):}

\medskip
\noindent
\texttt{Box path: [R, R, U, U, U, U, R]}\\
\texttt{(Validated: free behind-space each step)}
\end{promptbox}

\begin{promptbox}{Plan Agent \bad{} --- Failed Interpretation}{plancolor}{planborder}
\textbf{Input:}
\begin{itemize}
    \item \textbf{Task Description:} Use the Tool Agent’s box path to
    control the robot.
    \item \textbf{Tool Output:} \texttt{[R, R, U, U, U, U, R]}.
\end{itemize}

\textbf{Prompt (narrative as shown to the agent):}

The planning agent directly applies the box path to itself, failing to
understand that pushing a box is different from moving into a free
space.

\textbf{Response (Plan Agent thought and action list):}

\medskip
\noindent
\textbf{Thought:}\\
\emph{``Great, I'll execute these moves.''}

\medskip
\noindent
\textbf{Planned robot actions:}\\
\texttt{[R, R, U, U, U, U, R]}

\medskip
\noindent
\textbf{Outcome:} Blocked by walls; never gets behind the box to push
it. $\Rightarrow$ \textbf{Failure} \bad.
\end{promptbox}

\begin{promptbox}{Plan Agent \good{} --- Correct Interpretation}{plancolor}{planborder}
\textbf{Input:}
\begin{itemize}
    \item \textbf{Task Description:} Move the box from its start to the
    goal using the Tool Agent’s plan.
    \item \textbf{Tool Output:} \texttt{[R, R, U, U, U, U, R]}.
\end{itemize}

\textbf{Prompt (narrative as shown to the agent):}

The agent correctly translates the high-level plan (move the box) into
a sequence of agent-specific actions (get behind the box, then push).

\textbf{Response (Plan Agent thought process):}

\medskip
\noindent
\textbf{Thought Process:}
\begin{enumerate}
    \item Translate each box push to a required pre-push robot cell.
    \begin{itemize}
        \item Push Right $\rightarrow$ Robot must be Left of box.
        \item Push Up $\rightarrow$ Robot must be Below box.
        \item (And so on for L and D\ldots)
    \end{itemize}
    \item For each step in the box plan, route the agent to the
    corresponding pre-push cell (avoiding walls), then move one final
    tile in the push direction.
\end{enumerate}

\textbf{Output:} A valid sequence of robot moves that first routes
behind the box and then pushes it along the planned path, achieving the
goal \good.
\end{promptbox}

\subsection{MAS for Code}
\label{app:casestudy-code}

\paragraph{Code Generation Case Study.}
Task: Coder writes a solution; Unit-Tester writes tests.
Terminate=all tests pass. Otherwise: each agent revises its own previous
output using the environment feedback/results (Coder fixes code;
Unit-Tester fixes unit test), then re-run.

\paragraph{Before RL (\bad).}
Before RL: The Plan Agent gets a valid path for the box from Tool agent
but completely misses the point. It tries to follow the box's path
itself, runs straight into a wall, and fails instantly. It doesn't
understand its job is to push the box, not be the box.

\paragraph{After on-policy RL in MAS (\good).}
After on-policy RL in MAS: RL teaches the agent the difference. It
learns that rewards come from moving the box along the designated path.
This insight forces it to discover the correct low-level strategy:
first, navigate behind the box, then execute the push.

\begin{promptbox}{The Prompt (Unit-Test Generation)}{envcolor}{envborder}
\textbf{Input (environment prompt):}

A request is sent to a unit test generator to calculate the
``prettiness'' sum of an array \texttt{a} of size \texttt{n} using the
formula:
\[
\sum_{i=1}^{n} \sum_{j=1}^{n} \gcd(a_i, a_j)\cdot \gcd(i, j).
\]

The generator is instructed to produce the test input and output
following a specific format:
\begin{verbatim}
**Test Input:** input here
**Test Output:** output here
\end{verbatim}

\textbf{Output:} Problem specification and required I/O format for the
Unit-Tester.
\end{promptbox}

\begin{promptbox}{Unit-Tester \bad{} --- Failure 1: Missing Output}{testcolor}{testborder}
\textbf{Prompt (failure description):}

The generator only provided the test input, failing to include the
required output.

\textbf{Response (incorrect test case):}
\begin{verbatim}
**Test Input:**
n = 5
a = [3, 6, 2, 1, 4]

**Test Output:** (is missing)
\end{verbatim}

\textbf{Outcome:} The required \texttt{**Test Output:**} field is
missing entirely, so the test case is unusable \bad.
\end{promptbox}

\begin{promptbox}{Unit-Tester \bad{} --- Failure 2: Wrong Format}{testcolor}{testborder}
\textbf{Prompt (failure description):}

The output doesn't adhere to the specified \texttt{**Test...**} bolded
format.

\textbf{Response (incorrect test case):}
\begin{verbatim}
test_input:
n = 5
a = [3, 6, 2, 1, 4]

test_output:
77
\end{verbatim}

\textbf{Outcome:} Although both input and output are present, the format
violates the required bolded \texttt{**Test Input:**} and
\texttt{**Test Output:**} structure, so the checker rejects it \bad.
\end{promptbox}

\begin{promptbox}{Unit-Tester \good{} --- Success (MAS + AT-GRPO)}{testcolor}{testborder}
\textbf{Prompt (success description):}

Correct Case (MAS + AT-GRPO)\\
The output correctly follows the format, provides both input and
output, and ensures all variables are consistent.

\textbf{Response (correct test case):}
\begin{verbatim}
**Test Input:**
n = 5
a = [3, 6, 2, 1, 4]

**Test Output:**
77
\end{verbatim}

\textbf{Outcome:} The test fully respects the prescribed format and
contains consistent input and output fields, enabling reliable
automatic checking \good.
\end{promptbox}

\end{document}